\documentclass[5p]{elsarticle}
\usepackage{times}
\usepackage{epsfig}
\usepackage{graphicx}
\usepackage{amsmath}
\usepackage{multirow}
\usepackage{filecontents}
\usepackage{xcolor}
\usepackage{amssymb}
\usepackage[multi-part-units=single]{siunitx}
\usepackage{hyperref}
\newcommand{\cmmnt}[1]{\ignorespaces}
\journal{Medical Image Analysis}

\biboptions{authoryear}

\begin{document}

\begin{frontmatter}

\title{A Deep Learning Approach for Pose Estimation from Volumetric OCT Data}


\author[mymainaddress]{Nils Gessert\corref{cor1}}
\ead{nils.gessert@tuhh.de}

\author[mymainaddress]{Matthias Schl\"uter}
\ead{matthias.schluetert@tuhh.de}

\author[mymainaddress]{Alexander Schlaefer}
\ead{schlaefer@tuhh.de}

\address[mymainaddress]{Hamburg University of Technology, Schwarzenbergstra\ss{}e 95, 21073 Hamburg}
\cortext[cor1]{Corresponding author}

\begin{abstract}

Tracking the pose of instruments is a central problem in image-guided surgery. For microscopic scenarios, optical coherence tomography (OCT) is increasingly used as an imaging modality. OCT is suitable for accurate pose estimation due to its micrometer range resolution and volumetric field of view. However, OCT image processing is challenging due to speckle noise and reflection artifacts in addition to the images' 3D nature. We address pose estimation from OCT volume data with a new deep learning-based tracking framework. For this purpose, we design a new 3D convolutional neural network (CNN) architecture to directly predict the 6D pose of a small marker geometry from OCT volumes. We use a hexapod robot to automatically acquire labeled data points which we use to train 3D CNN architectures for multi-output regression. We use this setup to provide an in-depth analysis on deep learning-based pose estimation from volumes. Specifically, we demonstrate that exploiting volume information for pose estimation yields higher accuracy than relying on 2D representations with depth information. Supporting this observation, we provide quantitative and qualitative results that 3D CNNs effectively exploit the depth structure of marker objects. Regarding the deep learning aspect, we present efficient design principles for 3D CNNs, making use of insights from the 2D deep learning community. In particular, we present Inception3D as a new architecture which performs best for our application. We show that our deep learning approach reaches errors at our ground-truth label's resolution. We achieve a mean average error of $\SI{14.89 \pm 9.3}{\micro\metre}$ and $\SI{0.096 \pm 0.072}{\degree}$ for position and orientation learning, respectively.

\end{abstract}


\end{frontmatter}


\section{Introduction} \label{sec:introduction}

Tracking the pose of instruments and patients is a typical problem in many clinical scenarios, e.g., minimally invasive surgery (MIS) \citep{Bouget.2017} or transcranial magnetic stimulation \citep{Richter.2013}. Common commercially available optical and electromagnetic (EM) tracking systems reach an accuracy of \SIrange{0.2}{1}{\milli\metre} \citep{Kral.2013}. For optical tracking, a mean tracking error of \SI{0.22}{\milli\metre} has been achieved for clinical setups \citep{Elfring.2010}. EM tracking operates without a line of sight but generally reaches lower accuracy with a typical root mean square error (RMSE) of $\SI{1}{\milli\metre}$ \citep{Franz.2014}.  Some application scenarios in MIS require better accuracy, such as ophthalmic surgery, cochleostomy or neurosurgery. Moreover, the markers for optical tracking systems have a size of several centimeters which hinders application for these micro-scale scenarios. 

OCT represents a high-resolution image modality that is suitable for guiding microscale medical interventions. For example, OCT systems have been integrated into operating microscopes \citep{Lankenau.2007}, e.g., for ophthalmic surgery \citep{Tao.2014} and neurosurgery \citep{Finke.2012}. Moreover, OCT has been studied as a tracking system for cochleostomy by using artificial markers created with a laser \citep{Zhang.2014b}. The approach reached tracking accuracy in the micrometer range. These results motivate the use of OCT as a precise pose estimation and tracking system. 

Recently, deep learning-based frameworks have been applied for pose estimation problems. This includes methods to learn descriptors for 3D pose estimation from 2D images \citep{Wohlhart.2015} and full 6D pose estimation from RGB-D images \citep{Krull.2015}. Similarly, CNNs are considered a promising approach for surgical tool segmentation and pose estimation with recent successful applications \citep{Sahu.2016}. Taking a learning-based approach for pose estimation allows for independence from large markers which often comes at the cost of lower accuracy \citep{Bouget.2017}.

For OCT, tracking approaches have been proposed \citep{Laves.2017,Camino.2016}. However, these methods are limited to specific application scenarios such as skin or eye motion tracking using handcrafted features. Similar to pose estimation from time-of-flight camera images \citep{Krull.2015}, these approaches rely on 2D depth representations despite full volume data being available. In general, there are no deep learning approaches for OCT-based pose estimation so far.


For other medical image analysis task, such as segmentation of magnetic resonance imaging (MRI) data, 3D CNNs have been widely used \citep{Dou.2017,Havaei.2017,Kamnitsas.2017}. However, early 3D CNN architectures have been identified as lackluster due to simple architecture choices \citep{Yu.2017b} which leaves 3D CNN design as an open question. Moreover, to the best of our knowledge, 3D CNNs have not been applied to volumetric OCT data.

\begin{figure*}[htb]
	\centering
  		\centering
  		\includegraphics[scale=1]{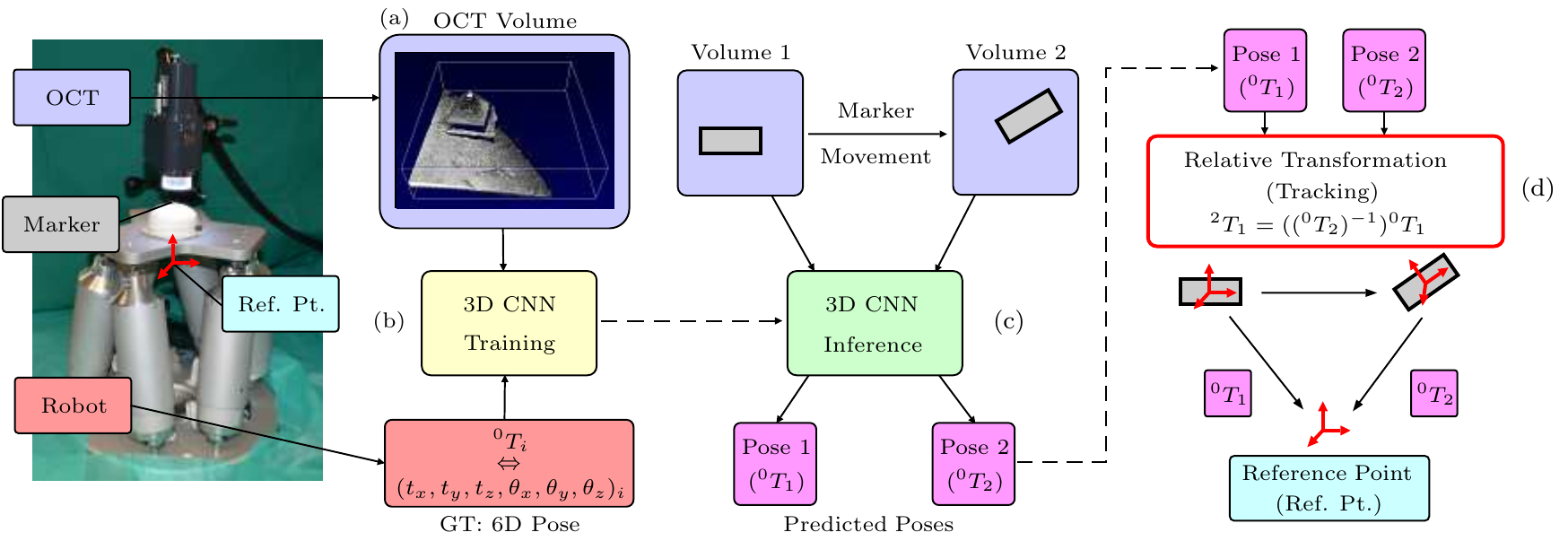}
  		\caption{Visualization of the approach. During the initial experiment, the hexapod robot moves the marker to predefined, randomly generated poses within the OCT's FOV. The poses are expressed with respect to the hexapod's reference point. In each pose, a volume is obtained while the robot is standing still (a). Next, the data samples are used for training a 3D CNN (b). Finally, the trained model (c) predicts a pose from an OCT volume. Two poses are used for tracking by obtaining the relative transformation (d) between the two marker poses.}
  		\label{fig:oct}
\end{figure*}

These considerations motivate a novel deep learning-based pose estimation approach for OCT. We take arbitrary small objects and turn them into a marker for pose estimation or tracking. To generate a training set, we acquire high-resolution volumetric OCT images of the object in different poses. We use a 3D CNN to learn highly accurate regression between volumetric images and object poses. Then, the 3D CNN can be used to estimate the object pose based on newly acquired volumetric images only. The object now acts as a marker that can be attached to surgical tools or patients to track their movement by inferring their pose changes from the marker. Figure~\ref{fig:oct} shows the data generation and tracking procedure in detail.

Our approach offers several advantages compared to the methods presented above. The marker's shape and size can be chosen arbitrarily, and it is easy to manufacture, e.g., with a 3D printer. A 3D CNN can be trained for any marker shape. This allows for adaptation of our framework to different clinical tracking scenarios with varying requirements. Moreover, compared to tool segmentation, our approach does not require sophisticated, manual labeling. Also, while having similar flexibility as a markerless approach, we benefit from the high accuracy of marker based systems as our 3D CNN is fitted to one specific geometry at a time.

In this paper we provide an in-depth analysis of our proposed method concerning its accuracy, the use of volumetric OCT data and 3D CNN architectures for pose learning with OCT volumes.

First of all, we address the fundamental question of tracking accuracy. We compare our novel deep learning-based pose estimation approach to a classic feature detection and registration-based method with a similar setup \citep{Zhang.2014b}.

Next, we motivate the use of volumetric data for deep learning-based pose estimation. We investigate how directly leveraging volume information with 3D CNNs compares to the typical use of 2D depth representations.

Regarding the choice of volume data as our image representation, we also analyze how 3D CNNs make use of the additional depth information. OCT is a modality that can provide deep, subsurface information. However, this depends on materials and whether they can be penetrated by infrared light. We investigate how subsurface information benefits 3D CNN learning by comparing markers with and without an identifiable inner structure. We provide quantitative accuracy results and qualitative saliency maps to show how 3D CNNs exploit volume information for pose estimation.

In order to show our method's robustness we also test our marker's performance when the OCT image is occluded. These results illustrate the performance of our method in practical scenarios where many new objects are likely to appear that have not been present during training.

Another aspect of our proposed framework is the deep learning model itself. As a part of our method, we extend 3D CNN usage to OCT volume data. Building 3D CNNs is not trivial since the models have larger numbers of parameters and high computational and memory requirements compared to 2D CNNs. We consider efficient CNN design principles such as Inception \citep{Szegedy.2016}, ResNet \citep{He.2016} and long-range feature transfer \citep{Ronneberger.2015,Yu.2017b} in order to build a new 3D CNN architecture called Inception3D. We compare it to several 3D CNN architectures for our pose estimation method and highlight how different design principles affect performance.


Summarized, the main contributions of this paper are as follows:

\begin{enumerate}
	\item We propose a novel deep learning method for direct pose estimation from volumes to track miniature markers with high accuracy.
	\item We show the advantages of a volume-based learning approach for pose estimation by comparing it to typical 2D depth-based tracking approaches.	
	\item We provide quantitative and qualitative evidence that 3D CNNs exploit the additional volume information well when using markers with internal features.
	\item Our work extends 3D CNNs to OCT volume data, and we introduce Inception3D as a new architecture for pose estimation and compare it to different CNN design principles.
\end{enumerate}

This paper is organized as follows. In Section~\ref{sec:relwork} we review related work. Then, we introduce our experimental setup, architectures, and methodology in Section~\ref{sec:methods}. We present results in Section~\ref{sec:results} and discuss them in Section~\ref{sec:discussion}. We draw final conclusions in Section~\ref{sec:conclusions}. 



\section{Related Work} \label{sec:relwork}

Our approach is linked to CNNs, pose estimation, and OCT imaging. 


\textbf{CNNs} have been widely used in various fields in computer vision such as classification \citep{Krizhevsky.2012}, object detection \citep{Girshick.2014}, pose estimation \citep{Toshev.2014} and semantic segmentation \citep{Long.2015}. Since their initial success in the ImageNet large scale visual recognition competition (ILSVRC2012), various new architectures and additions for CNNs have been introduced. The Inception architecture \citep{Szegedy.2015} showed success by utilizing different filter sizes on the same intermediate features in a network. This resembles the extraction of features at different scales. Residual connections were introduced to deal with the degradation problem in very deep networks \citep{He.2016}. These were also incorporated into a new iteration of the Inception architecture \citep{Szegedy.2017} that we use as a basis. \cite{Xie.2017} introduced ResNeXt, an architecture based on the ideas of Inception and residual learning. Their key contribution is the reduced number of hyperparameters that need to be chosen which makes the architecture easier to extend to new problems. \cite{Xie.2017} argue that sophisticated hyperparameter tuning hindered the application of successful architectures such as Inception to new domains. \cite{Li.2017} employed the Inception architecture on 3D data for 3D neuron reconstruction. However, the architecture was used with 2D kernels which leads to the CNN's kernels having 2D FOVs and thus no feature learning with volumetric data exploitation. 
Recently, the usage of 3D CNNs for volumetric MRI data was proposed. \cite{Dou.2016b} used 3D CNNs on MRI data for the detection of cerebral microbleeds. \cite{Brosch.2016} performed multiple sclerosis lesion segmentation on 3D MRI data. These approaches relied on simple CNN architectures and were therefore limited in their representation capability \cite{Yu.2017b}. Other approaches relied on custom 3D CNN designs, e.g. \cite{Havaei.2017} built a cascaded architecture and \cite{Dou.2017} relied on deep supervision with auxiliary classifiers \citep{Lee.2015} and dense output predictions. The U-Net design principle \citep{Ronneberger.2015} has been extended to 3D \citep{Cicek.2016} and is often found in 3D CNN architectures for segmentation tasks \citep{Litjens.2017}. The architecture is similar to an encoder-decoder scheme with feature propagation between similar resolution stages in the encoder and decoder part. \cite{Chen.2017b} built a related architecture with multi-scale feature aggregation at higher network levels. Moreover, \cite{Chen.2017} improved 3D CNN architectures by utilizing residual connections in a CNN for volumetric brain segmentation. \cite{Yu.2017b} refined this further by utilizing both short and long residual connections in a network. The latter is inspired by the feature propagation of U-Net. We also build on this idea, but we propagate information between different resolution stages instead of similar ones. So far, efficient design principles found in Inception and ResNet architectures have not seen a lot of attention for 3D medical image data although being successful in the 2D domain. Since these architectures are specifically designed for efficiency, we employ their design principles in the 3D domain where resources are often critical.

\textbf{Pose estimation} is a key problem in computer vision and has been widely studied and used in medicine. While typical approaches solve the task explicitly with known rigid body markers, machine learning-based approaches have gained popularity in clinical applications \citep{Bouget.2017}. In MIS environments, pose estimation is used for tracking of surgical tools or patients from endoscopic RGB videos. \cite{Allan.2014} performed tracking and 3D pose estimation of surgical tools from videos using linear Kalman filters. Recently, CNNs have been applied for the localization of tools in robot-assisted MIS surgery \citep{Sarikaya.2017}. Moreover, \cite{GarciaPerazaHerrera.2016} employ fully convolutional networks (FCN) for real-time segmentation and tracking of tools. Still, the application of CNNs in medical tracking tasks is rare, also due to the difficulty of obtaining large training sets \citep{Bouget.2017}.
 
In other fields, CNNs have been applied to pose estimation. CNNs have been used for pose estimation in RGB-D images. \cite{Wohlhart.2015} learned a semantic descriptor that separates image patches by object type and pose. Object recognition and pose estimation are performed by a nearest neighbor search which matches an image patch to a training sample based on their descriptors. The pose estimation is coarse and highly dependent on the density of training samples in the pose space. \cite{Krull.2015} took an analysis-by-synthesis approach for 6D pose estimation in RGB-D images. Rendered and observed image representations are fed as channels into a 2D CNN to predict an energy function value that is related to the target pose. \cite{Kehl.2016} employ CNNs in an unsupervised fashion on RGB-D patches for feature learning and subsequent 6D pose estimation. While images with a depth channel are frequently used, volumetric medical image data does not see usage for 6D pose estimation. We address this observation and show that directly using volumetric data is advantageous over the typical approach of relying on 2D depth representations.

\textbf{OCT} is an interferometric imaging modality with micrometer resolution and a typical field of view (FOV) of several millimeters range. OCT has been applied in surgical tasks through microscope integration, e.g., for ophthalmic surgery \citep{Ehlers.2014} and laser cochleostomy \citep{Zhang.2014b}. Also, OCT-based tracking setups fused with an RGB-D camera have been investigated \citep{Rajput.2016}. For laser cochleostomy, an OCT-based pose estimation framework has been proposed \cite{Zhang.2014b}. Artificial landmarks are applied to the patient's cochlea with a laser which are used for relative movement tracking. The high accuracy results imply the usability of OCT data for pose estimation and tracking. Moreover, tracking of a region of interest (ROI) has been performed with maximum intensity projections (MIPs) and handcrafted feature registration \citep{Laves.2017}. Again, this approach leverages 2D depth representations instead of full volumetric information.

Additionally, OCT image data has been recently used in conjunction with machine learning approaches for tasks not related to pose estimation. Segmentation of retinal fluids has been performed using CNNs with 2D OCT slices \cite{Schlegl.2015}. Moreover, tissue classification tasks have been addressed using recurrent neural networks \citep{Otte.2014} and CNN-based approaches \citep{Abdolmanafi.2017}. Also, detection of macular diseases has been addressed using CNNs \citep{Karri.2017,Lee.2017}. 

To the best of our knowledge, exploitation of volumetric OCT data with 3D CNNs has not been employed and is an open question for this imaging modality. We address this problem and compare different architectures that are new for the 3D CNN domain with our pose estimation method. Moreover, we address volumetric data exploitation of 3D CNNs and show its advantages over depth image-based pose estimation approaches found in the literature. 


\section{Methods} \label{sec:methods}

First, we introduce the setup for generating OCT and pose data. Second, the nature of our pose estimation framework is explained in detail. Third, the 3D CNN architectures we employ are introduced. 

\subsection{Data Generation and General Setup}

We employ a setup to automatically generate a set of image and pose data for learning. The setup consists of a hexapod robot, a spectral domain OCT (SD-OCT) device with a stand and a phantom to be used as a marker, see Figure~\ref{fig:oct}. The hexapod moves the marker inside the OCT's FOV and stops at predefined poses. The position part of the 6D poses is generated by randomly sampling positions in a 3D bounding box that covers the OCT's FOV size. Orientations are created by randomly generating rotation angles within an interval. All components are uniformly sampled from their respective space. The hexapod moves to a pose, stops, and an OCT volume is acquired. The volume is combined with the current pose to form a labeled data sample. This procedure is repeated several thousand times to create a dataset for training. As a result, our 3D CNNs receive an OCT volume containing the marker as their input and are trained in order to predict the pose with respect to the hexapod's reference point.

It should be noted that these labels require the models to implicitly learn the transformation between the hexapod reference frame and a marker coordinate frame. All poses are defined with respect to the hexapod. CNNs follow the universal function approximation theorem. Therefore, the complex model has the ability to learn the transformation. Moreover, this labeling approach allows fast, automatic data acquisition for large training sets. Also, the labeling strategy does not require pose estimation from images with a checker board, as typically used for learning-based pose estimation \citep{Brachmann.2014}.

Tracking is achieved by letting the CNN predict the marker's pose in two different volumes. Then, the relative transformation can be easily obtained by a matrix multiplication. This is depicted in the right part of Figure~\ref{fig:oct}. 

\subsubsection{OCT Imaging}

\begin{figure*}
  		\centering
  		\includegraphics[scale=0.5]{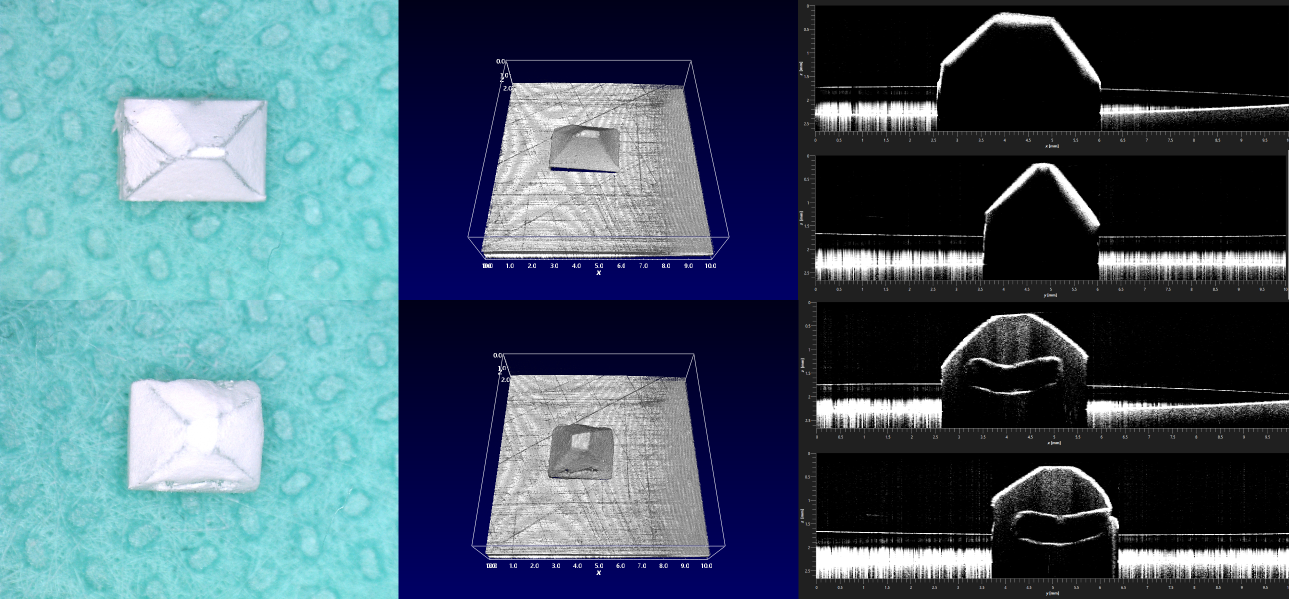}
  		\caption{The two markers we employ for training. Each row shows different image representations for each marker. From left to right: Digital microscopy image, rendered volume, B-Scan slices along the $x$ and $y$ direction. Note, that for the microscopy images the phantoms were coated for additional visibility which was not applied for the dataset acquisition. The first marker was milled from a polyoxymethylene (POM) block with a size of approximately $\SI{3.75 x 2.4 x 2}{\milli\metre}$. The second marker was 3D printed with Formlabs Resin with an approximate size of $\SI{3.2 x 2.68 x 1.9}{\milli\metre}$. The key difference is the inner structure of the second marker that is only visible in OCT volumes. We refer to the first marker as marker A and the second marker as marker B.}
  		\label{fig:markers}
\end{figure*}

The imaging device is an SD-OCT system which is based on interferometry. The technique's advantage is its high spatial resolution in micrometer range which makes it suitable for high accuracy tracking tasks. A broadband light source with a common center wavelength at \SI{1325}{\nano\meter} emits a beam that is split such that one part of it is directed at a reference mirror and the other part penetrates the object of interest. Light is scattered and reflected back and interferes with the reference signal part. A spectrometer captures the resulting interference spectrum that represents a 1D depth profile (A-scan) of the region of interest and is limited by the coherence length of the laser. Repeated scanning at different lateral positions results in a complete volume scan (C-scan) of the object of interest. The visibility of the object's interior structure largely depends on the object's reflective properties. If it reflects near infrared radiation very well, only the object's surface will be visible in an OCT volume. This is a very relevant property when considering the pose estimation task. Typical 6D pose estimation frameworks \citep{Krull.2015} also rely on surface information obtained with time-of-flight depth cameras. Therefore, it appears natural to employ a similar framework for OCT images if mostly surfaces are visible without internal features. We investigate this assumption by training both on volume data and 2D surface extractions. Also, we train both on an opaque marker, whose surfaces are hardly penetrated and a marker with a distinct inner structure, visible in OCT volumes. Both approaches provide insight on the importance of volume data usage. Figure~\ref{fig:markers} shows the different markers with the different properties. We refer to the opaque marker as marker A and the marker with an inner structure as marker B.

\subsubsection{Robot for Ground-Truth Annotation}

The hexapod robot shown in Figure~\ref{fig:oct} is used to move the marker within the OCT's FOV as well as for obtaining ground-truth 6D pose labels. Its pose is expressed with respect to a reference point slightly below its top plate. Translations relative to that point are denoted as $t_x$, $t_y$ and $t_z$. The rotations are expressed by rotation angles $\theta_x$, $\theta_y$, $\theta_z$ around each axis of a coordinate frame shifted by $t_x$, $t_y$ and $t_z$ from the reference point. Note, that rotations related to that point would lead to a translation of the phantom. Therefore, the center of rotation is shifted in $z$-direction to place it inside the OCT volume and minimize marker translations caused by rotations. A rotation matrix is expressed by consecutively rotating with $\theta_x$, $\theta_y$ and $\theta_z$ around the moving axes $x$, $y'$, $z''$, such that the rotation matrix can be expressed as $R = R(\theta_x)R(\theta_y)R(\theta_z)$. The rotation matrix $R$ and the translations are used to form a homogeneous transformation matrix that is used to obtain the relative transformation matrix as shown in the right part of Figure~\ref{fig:oct}. The target pose labels for learning take the form $p = (t_x,t_y,t_z,\theta_x,\theta_y,\theta_z)$. 

\subsection{3D CNN Architectures and Training Procedure}

Having obtained labeled data samples, the 3D CNN model can be set up, trained, optimized and used to predict poses. First, preprocessing steps are outlined where we set up datasets with 3D and 2D representations. Then, we described the novel 3D CNN architectures for 3D OCT images and explain design choices. 

\subsubsection{Preprocessing}

\textbf{For volume data}, the volume size needs to be adjusted first due to computational requirements. We downsample the volumes from the acquisition size of $128\times 128\times 512$ to $64\times 64\times 16$. The depth dimension is reduced with a larger factor than the lateral dimensions because its original pixel spacing is much smaller. As a result, the pixel spacing for each dimension of the volume represents the same cartesian distances. The target volume size is a trade-off between computational effort and potentially lost information during the downsampling process. The selected size leads to satisfactory results while keeping training times within feasible bounds. Note, that our pose estimation task does not allow us to perform subvolume sampling which is typically applied for large 3D input volumes \citep{Liefers.2017}. The pose is a global image property that would be lost in case of subsampling.
As a final preprocessing step, we subtract the training data set mean from each image to help gradient-based optimization \citep{Simonyan.2014}.

\textbf{For 2D depth data} representations we extract surface information from the OCT volumes to obtain a 2D depth representation that is similar to other RGB-D based 6D pose estimation frameworks \citep{Brachmann.2014}. This allows for comparison to other OCT-based tracking approaches where 2D depth representations were used for tracking a volume of interest with handcrafted feature matching \citep{Laves.2017}. 

We perform the extraction using MIPs from different views. This provides us with two different types of depth representations. The image index at which the maximum intensity was found represents the most intuitive notion of depth.

\begin{figure}
  		\centering
  		\includegraphics[scale=1]{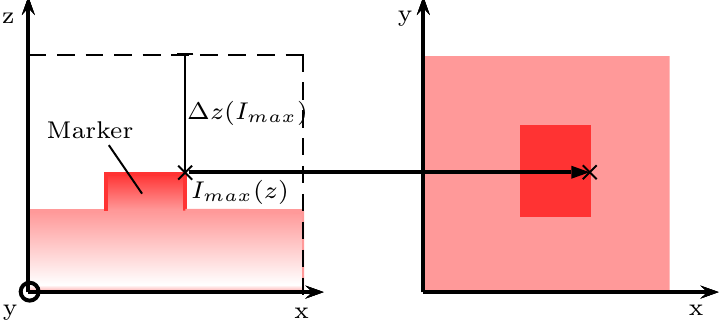}
  		\caption{The extraction process of 2D representations from an OCT volume. Here, the extraction for an $x$-$y$ projection is shown. Red color indicates the intensity in the volume. Typically, the highest intensity is found at the first surface hit by the infrared light. Inside, the intensity gradually decreases. Therefore, an MIP captures the surfaces visible in OCT data. For a depth map, the depth value $\Delta z(I_{max})$ is determined at every $x$-$y$ location and transferred to a 2D map. For an MIP, the intensity $I_{max}(z)$ itself is used at every $x$-$y$ location, as shown in the right part of the figure. Due to varying lighting properties along the $z$-direction, both methods capture depth in a 2D image.}
  		\label{fig:mip}
\end{figure}

However, the maximum intensity itself also provides depth information. Considering a curved Gaussian beam model of the OCT's infrared light, the intensity at the top of the volume (closer to the light source) will be different than at the bottom. Moreover, the MIPs can also carry rotation information as the back-scattering from surfaces changes based on the angle. Therefore, both the normalized depth index and the maximum intensities themselves are considered as 2D depth representations for learning. The extraction process is illustrated in Figure~\ref{fig:mip}. 
Since our data is volumetric, there are several options of which coordinate direction ($x$,$y$,$z$) should be chosen for extraction. Here, $x$ and $y$ are the lateral coordinate directions and $z$ is the depth direction along the OCT beam. Using several 2D projections from different angles is typically referred to as 2.5D and has been used for CNN training as a trade-off between less costly 2D and potentially richer 3D representations \citep{Roth.2016}. 

The straight forward choice is the use of the MIP along the z-direction as this is the actual travelling direction of the OCT light beam. Taking the maximum value along the $z$-direction results in a projection on the $x$-$y$ plane. Although this is the primary, relevant direction for OCT, some information is likely lost through the projection. To illustrate this, consider Figure~\ref{fig:mip}. Potentially useful information below the surface is lost entirely through projection. Therefore, we also include $z$-$y$ and $z$-$x$ projections in our datasets. To maintain spatial alignment, we perform the MIP extraction from a volume size of $64\times 64\times 64$. This results in five different 2D datasets that we compare to the volumetric dataset:
\begin{enumerate}
	\item $64\times 64 \times 1$ intensity values from the $x$-$y$ projection
	\item $64\times 64 \times 1$ normalized depth index values from the $x$-$y$ projection
	\item $64\times 64 \times 2$ normalized depth index values and intensity values from the $x$-$y$ projection
	\item $64\times 64 \times 3$ intensity values from the $x$-$y$, $z$-$x$ and $z$-$y$ projections
	\item $64\times 64 \times 3$ normalized depth index values from the $x$-$y$, $z$-$x$ and $z$-$y$ projections	
\end{enumerate}
The third dimension refers to the channel. 

In order to draw a connection between 2D and 3D data processing, we also consider the case of using 3D volume data with 2D kernels. Prior approaches handled OCT volume data by using 2D slices in the input data's channel dimension with 2D CNNs \cite{Schlegl.2015}. By default, a 2D kernel that is swept over a volume performs processing slice by slice without taking context between slices into account. For a meaningful comparison to 3D CNNs, we extend the 3D volumes by a channel dimension for 2D kernel processing. Each channel contains a shifted version of the volume along the $z$-direction. Therefore, when processing each slide with a 2D kernel, the neighboring slices are also taken into account.

Summarized, we use five datasets with 2D depth representations for comparison to a volumetric dataset. This provides a comparison on how computationally cheaper 2D representations perform against more costly 3D data when being trained with a 2D CNN and 3D CNN, respectively. The baseline dataset for our evaluation is the volumetric dataset. 

\subsubsection{3D CNN Architectures} \label{sec:arch}

First, we motivate our general 3D CNN approach for the 6D pose estimation task at hand. Then, we describe the different architectures we employ with the respective design principles we followed for their construction.

Although CNNs have been popular for several years, application to volumetric input data in medical imaging is still rare \cite{Greenspan.2016} and to our knowledge not available at all for OCT volume data. Therefore, our architecture follows popular design choices from the deep learning community for 2D applications and also considers successful approaches on MRI volume data. 

\begin{figure*}
  		\centering
  		\includegraphics[scale=1]{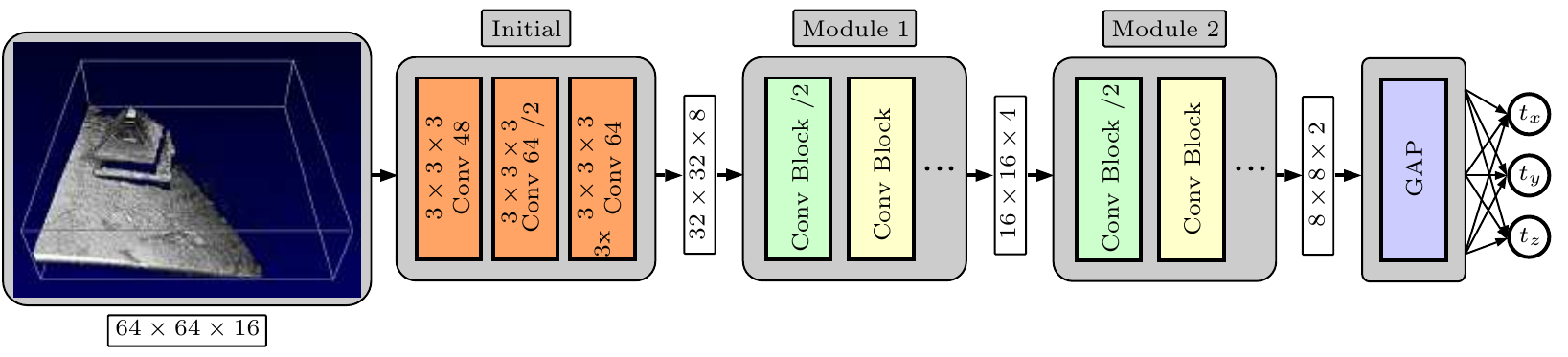}
  		\caption{The generic architecture we propose for our framework. The initial part, intermediate volume sizes, and the output part are identical for every architecture. The modules are individually designed for each specific architecture. All modules start with a convolutional block that reduces the spatial input dimension by half with a stride of two.  Note, that the network's output size is three because we train one model each for position and orientation learning. Here, the output for position learning is shown.}
  		\label{fig:genarch}
\end{figure*}

The complete 3D CNN consists of several convolutional layers which represent a feature extraction stage and an output layer for the regression itself. The convolutional layers consist of a set of 3D kernels that are swept over the input and create several output feature volumes. The 3D property of the kernels leads to volumetric receptive fields which enable volume information exploitation.

Our principle network design is shown in Figure~\ref{fig:genarch}. After the volumetric input, some initial layers follow, which are identical for all architectures we build. Immediately after the first layer, we halve the input's spatial dimension. We employ convolutional layers with stride two instead of the typical max pooling layer, following the idea of simplistic design \citep{Springenberg.2014}. Then, groups of architecture-specific layers follow, which we refer to as modules. At the module input, the first layer always reduces the input size by half in all spatial dimensions. Every architecture comes with two modules, representing our main feature extraction stage with the most model parameters and the largest influence on performance. After two modules, we apply global average pooling to reduce the current feature volume to a feature vector. This approach acts as a regularization as the following fully-connected layer has significantly fewer parameters \citep{Lin.2013}. The feature vector is fed into the output layer that predicts the pose as continuous regression. We chose to train separate networks for position and orientation. Therefore the CNN output is always a vector with three elements. We motivate this choice when describing the target vectors in detail in Section~\ref{sec:traincnn}. We compare this approach to direct prediction of the entire pose vector.

The general architecture focuses on feature extraction at intermediate volume sizes of $16 \times 16 \times 4$ and $8 \times 8 \times 2$. Note, that the volumes are padded to retain the desired volume sizes after convolutions. Considering the spatial dimension of the $z$-axis, moving these main extraction stages to smaller volumes is not reasonable. Shifting the main extraction towards larger volumes is suboptimal as well since computational effort would increase tremendously. 

For the modules in Figure~\ref{fig:genarch} we employ different types of architectures to highlight the advantage of our network design. Each model introduces a different additional property that leads to our design of Inception3D, the main architecture we introduce in this paper. To maintain a fair comparison, we try to keep the architectures similar with respect to the number of parameters (4 million) and features learned.

To keep architecture design straight forward, we follow previous design principles for the 2D domain. \cite{Simonyan.2014} showed that smaller kernel sizes are preferable for CNNs which is why we only employ $3\times 3\times 3$ filters for feature learning and $1\times 1\times 1$ filters for changing feature map sizes. Moreover, we increase the number of feature maps in our modules each time the spatial feature dimensions are halved.

Additionally, we employ batch normalization before every activation to reduce covariate shift \cite{Ioffe.2015}. The activation functions are of type ReLu \cite{Glorot.2011}. 

\begin{figure}
  		\centering
  		\includegraphics[scale=1]{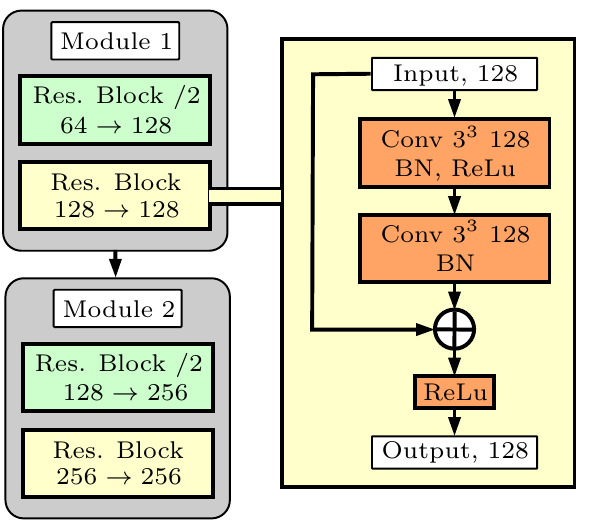}
  		\caption{The architecture of the ResNetA3D model is shown. Each module contains two residual blocks where the first block in each module reduces the spatial dimension by half and increases the feature map dimension by a factor of two. Conv $3^3$ indicates a filter size of $3\times 3\times 3$. Note, that the ReLu activation is applied after the addition. The residual blocks follow the concept of \citep{He.2016} and have been employed in 3D CNNs by \citep{Yu.2017b}. We see this architecture as state-of-the-art for 3D CNNs that follow successful 2D CNN architectures.}
  		\label{fig:resnetold}
\end{figure}

\textbf{ResNetA3D} is an architecture that we base on current state-of-the-art 3D segmentation CNNs such as \citep{Chen.2017,Yu.2017b} to provide a meaningful comparison to our other models. Several blocks of this architecture are joined to modules as shown in Figure~\ref{fig:resnetold}. The key feature of this architecture compared to plain convolutional blocks is the use of residual connections \citep{He.2016}. The idea of this concept is to learn a residual $\mathcal{F}(x) = \mathcal{H}(x)-x$ instead of the desired mapping $\mathcal{H}(x)$ where $x$ is the block's input. Residual connections are frequently used in the 2D image domain with numerous variations \citep{Szegedy.2017,Zagoruyko.2016} and recently the concept was employed for 3D prostate segmentation \citep{Chen.2017}. Therefore, we see this model as a baseline architecture reflecting the application of 2D design principles in the 3D image domain. Note, that this model is expensive regarding its number of parameters as is does not employ downsampling in the number of feature maps which is introduced next. Therefore, the network comes with a smaller depth to maintain a similar amount of parameters.

\begin{figure}
  		\centering
  		\includegraphics[scale=1]{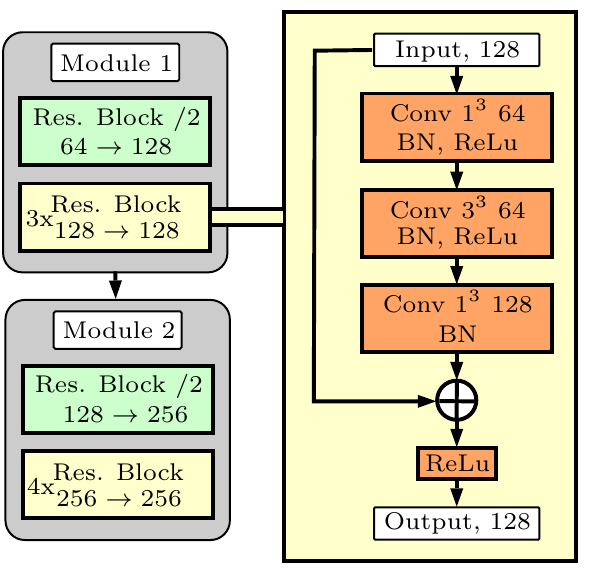}
  		\caption{The Architecture of the ResNetB3D model is shown. Each module contains four and five residual blocks respectively where the first block in each module reduces the spatial dimension by half and increases the feature map dimension by a factor of two. Conv $1^3$ and $3^3$ indicate filters sizes of $1\times 1\times 1$ and $3\times 3\times 3$, respectively. The residual blocks follow the concept of \citep{He.2016} and introduce downsampling for the feature map dimension which significantly reduces the number of parameters and computational effort. This enables a deeper architecture compared to model ResNetA3D.}
  		\label{fig:resnetnew}
\end{figure}

\textbf{ResNetB3D} is a model that extends the concept of residual blocks from ResNetA3D by adding $1 \times 1 \times 1$ convolutions for downsampling and upsampling in the feature map dimension, as shown in Figure~\ref{fig:resnetnew}. Often, this idea is described as a \textit{bottleneck}. Furthermore, the method should be distinguished from \textit{spatial} downsampling which acts on the images' width, height and depth and helps to increase the implicit receptive fields. Reducing the feature map dimension follows the idea of dimensionality reduction which assumes that most of the input's information can be preserved in a lower dimensional embedding. This concept was also used in the original 2D ResNet architecture \citep{He.2016}. However, to our knowledge, it has not been employed for 3D CNN learning tasks. This concept is particularly important for costly 3D CNNs as this method reduces the number of parameters and computational effort for the model. Note, that this design principle allows for a deeper model with more layers than ResNetA3D. 

\begin{figure}
  		\centering
  		\includegraphics[scale=1]{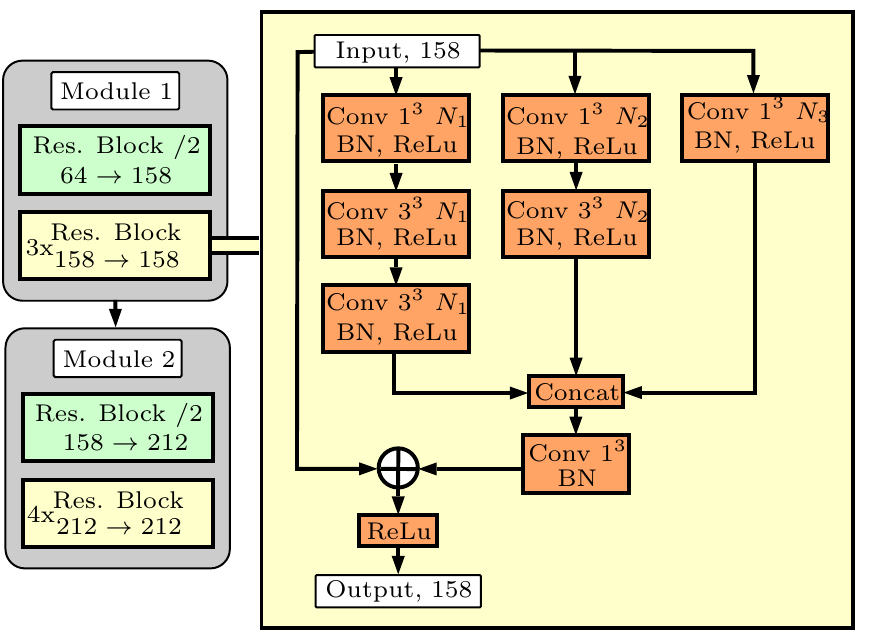}
  		\caption{The architecture of the Inception3D model is shown. Each module contains four and five residual blocks respectively where the first block in each module reduces the spatial dimension by half and increases the feature map dimension by a factor of two. Conv $1^3$ and $3^3$ indicate filter sizes of $1\times 1\times 1$ and $3\times 3\times 3$, respectively. The inception blocks follow the concept of \citep{Szegedy.2017} and introduce multiple paths in each residual block in addition to feature map downsampling. Note, that the residual part of each block is scaled by $s=0.2$ as suggested by \cite{Szegedy.2017}. The parameters $N_i$ are shown in Table~\ref{tab:inceptparams} as they are individually chosen for each block and path. The final $1\times 1\times 1$ convolution in each inception block recovers the original feature map size $N_M = \sum_{i} N_i$.}
  		\label{fig:inception3D}
\end{figure}

\begin{table}
	\centering
	\begin{tabular}{l l l l}
		& $N_1$ & $N_2$ & $N_3$ \\ \hline \\
		Module 1 Res. Block /2 & $64$ & $64$ & $30$ \\
		Module 1 Res. Block & $42$ & $42$ & $20$ \\
		Module 2 Res. Block /2 & $86$ & $86$ & $40$ \\
		Module 2 Res. Block & $64$ & $64$ & $30$ \\
	\end{tabular} \\
	\caption{Parameter choices for the residual blocks of the inception architecture, see Figure~\ref{fig:inception3D}}
	\label{tab:inceptparams}
\end{table}

We propose \textbf{Inception3D} as a new 3D CNN architecture which is inspired by Inception-ResNet \citep{Szegedy.2017}. We make use of the previous models' properties and additionally introduce the concept of multi-path convolutional blocks, as shown in Figure~\ref{fig:inception3D}. The individual parameter choices for the convolutional layer sizes are shown in Table~\ref{tab:inceptparams}. The multi-path approach is motivated by the idea of feature extraction at different scales which is expected to yield more representative features \citep{Szegedy.2015}. Note, that this architecture is difficult to design, in particular, as more design choices need to be made. We address this problem by simplifying Inception3D without taking away its core concepts. Compared to \cite{Szegedy.2017}, we employ a single type of Inception module with the same number of feature maps (width) for all filters in each path. Compared to our other models, we individually choose each block's width, and we augment the architecture with long-range residual connections.

\begin{figure}
  		\centering
  		\includegraphics[scale=1]{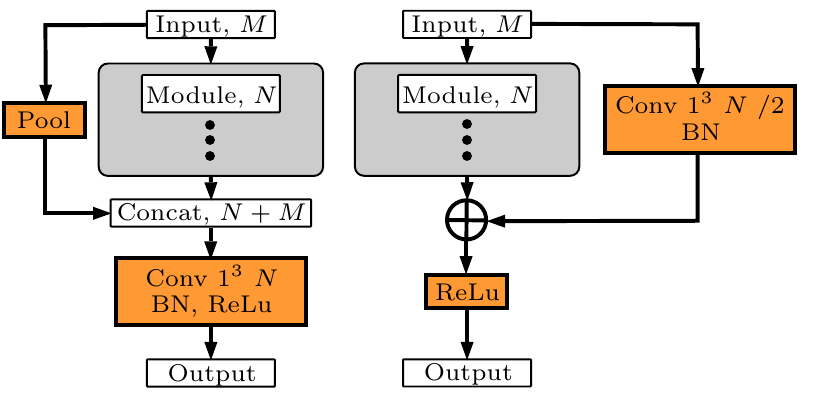}
  		\caption{Two types of long range connections over the modules of Inception3D are shown. Left, the transfer of features between stages is shown with a concatenation of features from different levels. Right, feature transfer through a long range residual connection is shown. $M$ denotes the number of input feature maps, $N$ the number of output feature maps from the module. Pool indicates $2\times 2\times 2$ average pooling to match the module's spatial dimensionality reduction. Conv $1^3$ indicates $1\times 1\times 1$ convolutions for adjustment of the number of feature maps. For the residual connection, the convolution is applied with a stride of two to match the module's spatial dimensionality reduction.}
  		\label{fig:rescon}
\end{figure}

The idea of long-range residual connections is inspired by \cite{Yu.2017b} where connections between the same feature map stages are applied in a U-net-like \citep{Ronneberger.2015} encoder-decoder network. We extend this idea by transferring features between different feature map scales. For comparison, we also use the original idea of U-net for feature transfer \citep{Ronneberger.2015}. While residual connections perform an addition operation when features are fused, U-net concatenates the features to a larger feature map. For the latter, we perform a subsequent $1\times 1\times 1$ convolution that reduces the feature map size back to the original size after concatenation. In this way, the network can learn which combination of high- and low-level features is needed. The idea behind this approach is that pose estimation requires both local and global features. The latter are necessary for the object's general position in the image while the former allow for fine-grained distinction of similar poses. Both skip connection approaches are shown in Figure~\ref{fig:rescon}. 


\begin{figure}
  		\centering
  		\includegraphics[scale=1]{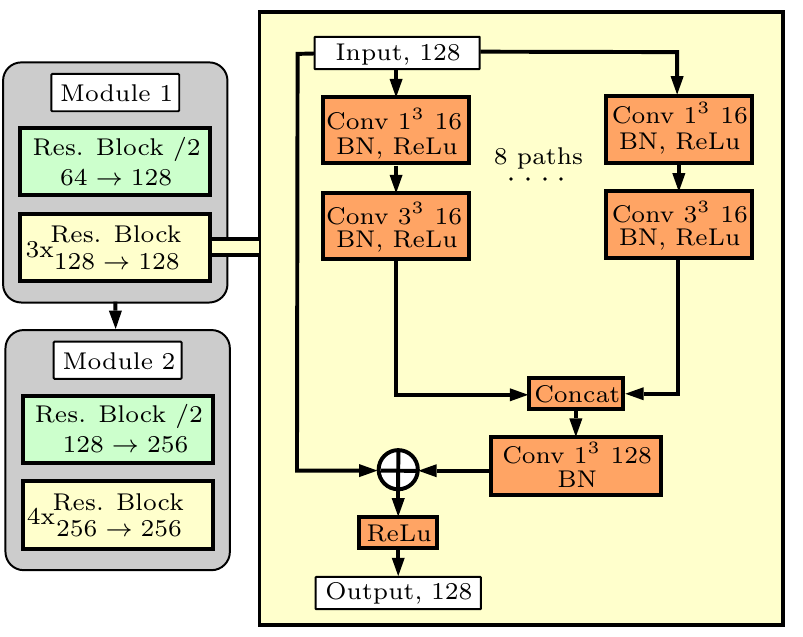}
  		\caption{The architecture of the ResNeXt3D model is shown. Each module contains four and five residual blocks, respectively where the first block in each module reduces the spatial dimension by half and increases the feature map dimension by a factor of two. Conv $1^3$ and $3^3$ indicate filters sizes of $1\times 1\times 1$ and $3\times 3\times 3$, respectively. The residual blocks follow the concept of \citep{Xie.2017} and provide a more simplified version of inception as all paths are identical. This significantly reduces design efforts when constructing a new architecture for a new application. Note, that ResNeXt architecture can be implemented differently, e.g., with grouped convolutions \citep{Xie.2017}.}
  		\label{fig:resnext}
\end{figure}

\textbf{ResNeXt3D} is similar to the Inception idea with a multipath architecture which is inspired by \citep{Xie.2017}, see Figure~\ref{fig:resnext}. The key idea is to utilize all of the above models' ideas with simplified design principles. The multiple paths idea from Inception is adopted by splitting up the single convolution path from ResNetB3D. The number of paths is referred to as \textit{cardinality} which is considered the key hyperparameter to choose for this type of architecture \citep{Xie.2017}. The resulting architecture is easy to tune as all paths are identical compared to Inception, where each path is carefully tuned individually. Therefore, the key difference between ResNeXt3D and Inception3D is simpler architecture design for the former. 


All in all, we propose four different architectures for the 3D image domain. Inception3D is our main architecture which we compare to the different design principles of our other models. ResNetA3D is a baseline with residual blocks that are found in typical 3D CNNs \citep{Yu.2017b}. For ResNetB3D we introduce the use of downsampling in the feature map dimension for more effective feature representation with the same amount of parameters. We augment Inception3D, our main architecture, further with multi-path blocks and long-range residual connections for optimal performance. Lastly, ResNeXt3D shows how a network with little design effort compares to our similar but carefully tuned Inception3D architecture. These architectures highlight how different design principles affect performance for our pose estimation method. A summary of all architectures is shown in Table~\ref{tab:allarch}. Also note, that all our architectures are very efficient in terms of the number of parameters. For comparison, the standard ResNet50 architecture \citep{He.2016} with 16 residual blocks and 2D convolutions comes with 21 million parameters. Inception-ResNet \citep{Szegedy.2017} contains 22 blocks and 56 million parameters.

\begin{table*}
	\centering
	\begin{tabular}{l l l l l}
		& ResNetA3D & ResNetB3D & Inception3D & ResNeXt3D \\ \hline \\
		Residual Connections & Yes & Yes & Yes & Yes  \\
		Bottleneck & No & Yes & Yes & Yes \\
		Multi-Path & No & No & Yes & Yes \\
		Individual Path Design & No & No & Yes & No \\
		\# of Parameters & \num{6161907} & \num{3451507} & \num{3568913} & \num{3042931} \\
		\# of Blocks & 4 & 9 & 9 & 9 \\
	\end{tabular} \\
	\caption{Overview of the different architectures we employ for pose estimation. All models employ residual connections. Except for ResNetA3D, all models make use of downsampling in the feature map dimension, i.e., the bottleneck principle. Inception3D and ResNeXt3D additionally contain multiple paths at each stage, representing feature extraction at different scales. Inception3D's pathes are individually fine tuned while ResNeXt3D follows simple design rules for its path design. Lastly, the total number of parameters and blocks is provided for each model. Note, that ResNetA3D only has 4 blocks in order to keep its number of parameters in a similar range.}
	\label{tab:allarch}
\end{table*}

\subsubsection{Training the 3D CNNs} \label{sec:traincnn}

The learning task is formulated as a regression problem, which is why the error function to be minimized is chosen to be the mean squared error (MSE) between network outputs and ground-truth labels. We define the MSE as

\begin{equation}
	MSE = \frac{1}{d}\sum_{i=1}^{d}\frac{1}{N_B}\sum_{j=1}^{N_B}(y_i^{j}-\hat{y}_i^{j})^2
\end{equation}

where $d$ is the number of outputs, $N_B$ the batch size, $y$ the ground-truth label and $\hat{y}$ the network's predictions. The CNNs are trained with mini-batch gradient descent. We use the Adam algorithm \citep{Kingma.2014} as a state-of-the-art optimizer with an initial learning rate of $l_r = \num{e-4}$. When the validation error saturates, the learning rate is reduced by a factor of $5$ until we observe no further improvement. The decay rates for the first and second order statistical moment estimates are chosen according to \cite{Kingma.2014} with $\beta_1=0.9$ and $\beta_2=0.999$. Similarly, the decay rate for the moving average in batch normalization layers is chosen to be $\beta = 0.9$. Following \cite{Ioffe.2015}, we do not apply other regularization methods.

We split the data set into training, validation and test sets. The validation set is used for fine-tuning hyperparameters, the test set is used for evaluating the final performance. During training, we use a batch size of $N_B = 15$.

The labels used for training are provided by the hexapod robot. Due to the OCT's limited FOV, the positions are limited to $t_x,t_y \in [\SI{-5}{\milli\metre},\SI{5}{\milli\metre}]$ and $t_z \in [\SI{-1.2}{\milli\metre},\SI{1.2}{\milli\metre}]$. Similarly, we limit rotations to $\theta_x,\theta_y,\theta_z \in [\SI{-10}{\degree},\SI{10}{\degree}]$. For training, we rescale the regression outputs to a range of $[0,1]$. In particular, we rescale every output component $y_i^j$ individually to a range based on the training set. The scaled outputs $\tilde{y}_i^j$ are defined as

\begin{equation}
	\tilde{y}_i^j = \frac{y_i^j - y^{min}_i}{y^{max}_i - y^{min}_i}
\end{equation}

where $y^{min}_i$ and $y^{max}_i$ are the minimum and maximum value of output $y_i$ in the training set. For evaluation we transform the network's predictions $\hat{y}$ back to the original scale and calculate error metrics on those values.

Another question that we address is whether training a single CNN for the entire pose label is the optimal choice. Multi-output regression has been addressed both by training a single model for the entire output and by training individual models for each output \citep{Borchani.2015}. We study three different approaches. First, we train a single CNN to predict the complete 6D pose. Second, we train one CNN each for position and orientation prediction. Third, we train one CNN each to predict a single component of the pose vector. We choose the best performing approach for all other experiments.

\subsubsection{Visualizing What CNNs Learn} \label{sec:saliency}

Understanding and visualizing what CNNs learned after training is an important issue in the field of deep learning \citep{Simonyan.2013}. In particular, for the problem at hand, it is crucial to understand what kind of image properties the CNNs leverage for pose estimation. In general, CNNs for classification are either visualized by image generation through maximization neuron activations or with saliency maps \citep{Zeiler.2014}. We utilize the latter since activation maximization is not immediately applicable to regression with continuous output values. Saliency maps visualize which region in a particular input image has the largest influence on a certain activation in the network. This is achieved by computing the partial derivative of the activation with respect to the current input image, leading to a gradient image

\begin{equation}
	S_{x,y,z} = \sum_{i} \frac{\partial y_i}{\partial I_{x,y,z}}
\end{equation}

where $S_{x,y,z}$ is the saliency map, $I_{x,y,z}$ is the input image, and $y$ is a vector of activations. The partial derivatives for each vector element $y_i$ are summed up to form the saliency map. We set $y$ to be the output of our network, and thus, a saliency map tells us which region of an image leads to the largest change in the output. This allows us to visualize what our CNN focuses on when being trained on 2D data, when being trained on the marker with a surface structure and when being trained on a marker with inner features.

To enhance the saliency maps, we utilize guided backpropagation \citep{Springenberg.2014}. The key idea of this approach is to combine normal backpropagation with the deconvolution idea of \cite{Zeiler.2014}. Effectively, guided backpropagation changes the backward pass of the ReLu activation function such that negative gradients and thus components that reduce the target activation are suppressed. The method has been shown to perform better than normal backpropagation and deconvolutional visualization, for details see \citep{Springenberg.2014}.

All in all, we support our investigation of depth exploitation in volume data for our 6D pose estimation technique by providing an intuitive visualization of what the CNNs learn.

\subsubsection{Online Pose Estimation and Robustness Towards Occlusion} \label{sec:occl}

In order to show our method's potential for clinical application scenarios, we also investigate the CNNs' inference runtime and their robustness towards occlusion in the OCT volumes. 

We compare inference runtimes for three different approaches. First, we use Inception3D which employs 3D convolutions and processes volume data. Second, we use a 2D variant of Inception3D with 2D convolutions for the 2D depth representations. Third, we use the 2D variant to process volume data as slices. We investigate whether the different mathematical operations and input data lead to differences in processing time. 

We measure the time that passes between feeding a single input to the model and receiving the respective output. We provide mean and standard deviation for 100 single input passes to the model.

\begin{figure*}
  		\centering
  		\includegraphics[scale=0.22]{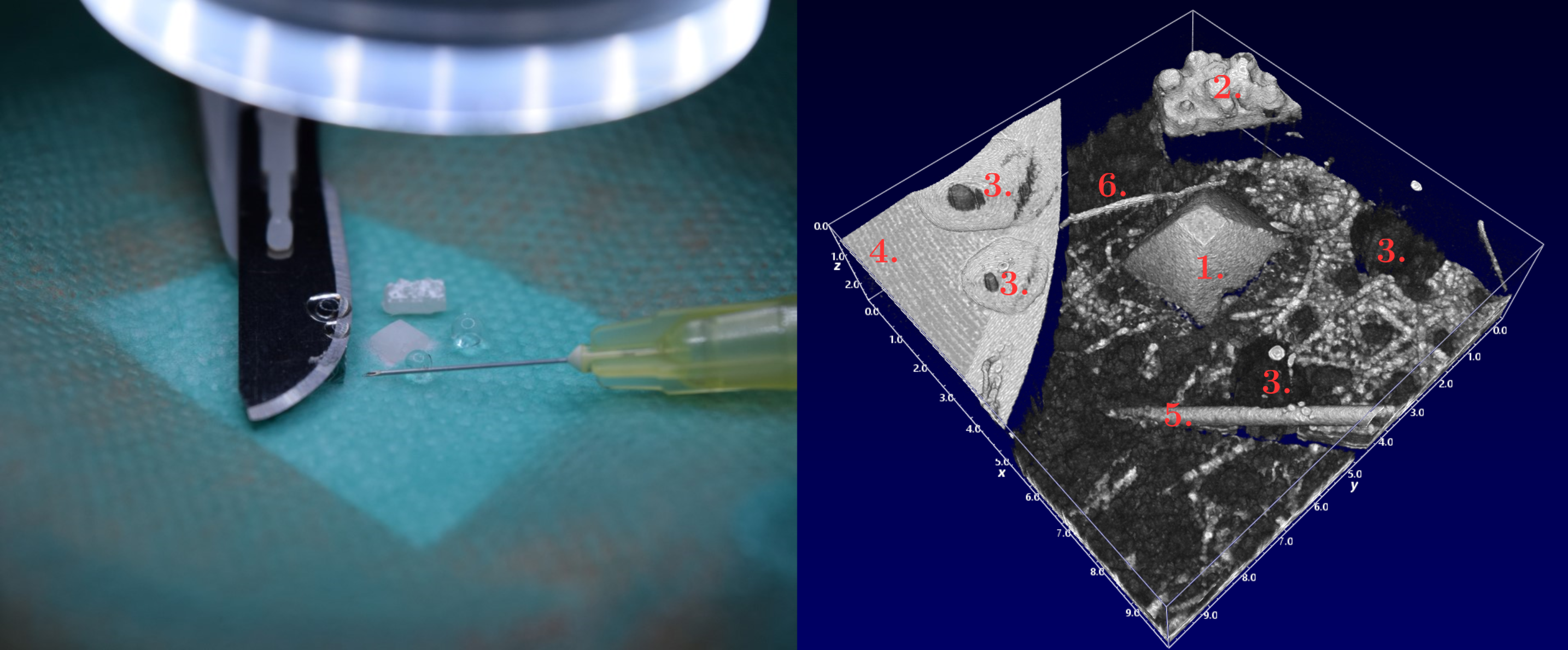}
  		\caption{Example for the occlusion dataset. Left, a photography of the setup is shown. Right, the corresponding OCT volume is shown. We use marker B for this experiment. Note, that we vary both the position of the objects and the objects themselves during data acquisition of this set. 1. marker B 2. printed geometry/arbitrary marker 3. water droplets 4. scalpel 5. needle of a syringe 6. cloth fibre.}
  		\label{fig:occlusion}
\end{figure*}

Furthermore, we investigate how our models react to occlusion in OCT volume data. For this purpose, we acquired an additional dataset where we added random objects around the marker. The occluding objects were repositioned and changed during training. We used a variety of objects with different reflective properties such as a scalpel, parts of a syringe, needles, cloths, different plastic and metal parts, surgical scissors, printed geometries that could be used as markers and water droplets on top of and next to the marker. An example occlusion scenario is shown in Figure~\ref{fig:occlusion}. Our marker is the only object constantly appearing in all volumes, and we investigate whether this helps the model to learn robustness towards all other objects.

For testing we split off a dataset that contains objects that are not present anywhere else in the training dataset. Therefore, performance on this test set indicates how well the model deals with objects that it has never seen before. This provides a realistic impression on how the model will perform in practice where new objects are likely to appear in the OCT volumes.

\section{Results} \label{sec:results}

In this section, we present our results. First, we introduce our acquired datasets and the experimental setup. Second, we provide a description of our evaluation strategy. Third, we provide the results themselves.

\subsection{Experimental Setup and Data} \label{sec:data}

Marker A was milled from a block of polyoxymethylene (POM) with an asymmetric prism shape, see Figure~\ref{fig:markers}. The material reflects the infrared light very well, which is why mostly its surface is visible in an OCT volume, not its interior. The second marker was 3D printed with Formlabs Resin to obtain an inner structure. For both markers we acquired several thousand data samples each, using roughly $\SI{80}{\percent}$ for training and $\SI{10}{\percent}$ for validation and $\SI{10}{\percent}$ for testing. Additionally, we acquired a dataset that contains occlusions as described in Section~\ref{sec:occl}. Note, that there is no validation set for the occlusion dataset as we directly use it with our models that were fine-tuned on the other two datasets. An overview of the datasets is shown in Table~\ref{tab:datasets}. All results we present refer to the test sets.

\begin{table}
	\centering
	\begin{tabular}{l l l l}
		& Marker A & Marker B & Occlusion \\ \hline \\
		Training & 5850 & 5850 & 15000 \\
		Validation & 900 & 900 & - \\
		Testing & 900 & 900 & 2875 \\
	\end{tabular} \\
	\caption{Number of samples for each dataset. The occlusion dataset was recorded with marker B.}
	\label{tab:datasets}
\end{table}

The OCT device is a Thorlabs Telesto I SD-OCT. Its lateral resolution is $\SI{15}{\micro\metre}$ and its depth resolution is $\SI{7.5}{\micro\metre}$. Its FOV covers a volume of $\SI{10 x 10 x 2.66}{\milli\metre}$. Volume images are acquired with a size of $\num{128 x 128 x 512}$ voxels. In the setup shown in Figure~\ref{fig:oct} only the OCT's scan head is visible.

The robot is a 6-axis H-820.D1 hexapod distributed by Physik Instrumente GmbH. It allows travel ranges of $\SI{20}{\milli\metre}$ for translations and $\SI{15}{\degree}$ for rotations, covering the OCT's FOV. Regarding accuracy, the robot is limited by a translational repeatability of $\pm\SI{20}{\micro\metre}$ and a rotational repeatability of $\pm\SI{11.46e-3}{\degree}$. The range of positions covered by the hexapod robot in the experiment corresponds to the OCT's FOV. The rotations are limited to a range of $(\SI{-10}{\degree},\SI{10}{\degree})$ for each axis.

The 3D CNN implementation leverages the TensorFlow environment \citep{Abadi.2016}  and training is performed with graphics cards of type nVidia GTX 1080 Ti with 11GB VRAM. 

\subsection{Evaluation Strategy}

We provide the results of the analysis of our pose estimation method in several steps: 

\begin{enumerate}
	\item We show general accuracy results and motivate the use of deep learning by comparing our framework to a more classic approach. For this comparison we use our best performing model Inception3D and the best performing marker B. Moreover, we show results for our choice of splitting position and orientation learning.
	\item We show pose estimation accuracy for 2D depth representations for 2D CNN training and 3D volumes for both 2D and 3D CNN training. Again, we employ Inception3D with a 2D counterpart for this comparison. We use marker A for this comparison. The marker is best suited for comparison with 2D depth representations as it largely shows surface information in OCT volumes.
	\item We show how marker A compares to marker B in order to highlight the effects of inner marker structure for 3D CNN learning. We use Inception3D for this comparison.
	\item We visualize what our 3D CNN learns using saliency maps as described in Section~\ref{sec:saliency}. This adds qualitative results and a better understanding for the previous, quantitative results.
	\item We show the suitability of our method for online pose estimation by providing inference times for 2D and 3D CNN data processing.
	\item We show our method's robustness by using our Inception3D model for a dataset with heavy occlusion.	
	\item We compare the 3D CNN models introduced in Section~\ref{sec:arch} with respect to their performance for our pose estimation method. We use both markers for this comparison.

\end{enumerate}

We evaluate pose estimation accuracy using the mean absolute error (MAE), relative MAE (rMAE) and average correlation coefficient (aCC) which are typical measures for regression tasks \citep{Borchani.2015}. The relative MAE is obtained by dividing the MAE by the ground-truth label's standard deviation. All reported accuracy values are derived from the independent test sets.

\subsection{Pose Estimation Accuracy}

\begin{figure}
	\centering
	\includegraphics[scale=1]{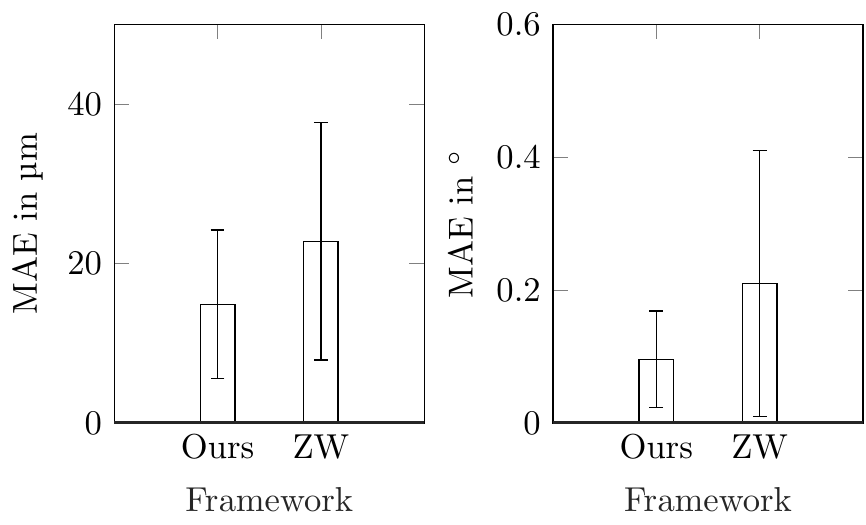}
	\caption{The comparison of test errors of our framework (Ours) with the approach of \cite{Zhang.2014b} (ZW) is shown. Left, the position MAE (with standard deviation) in micrometer is shown. Right, the orientation MAE (with standard deviation) in degree is shown. We used marker B for these results.}
	\label{fig:compare}
\end{figure}

First, we show how the use of a deep learning technique for 6D pose estimation from volume data compares to a classic feature based method. For the comparison, we use the related framework of \cite{Zhang.2014b}. Their method is similar to ours in terms of the experimental setup as they use OCT as an imaging modality and a hexapod for movement.

The comparison is shown in Figure~\ref{fig:compare}. Our approach outperforms the other framework with an MAE of $\SI{14.89 \pm 9.3}{\micro\metre}$ for our method compared to $\SI{22.8 \pm 14.9}{\micro\metre}$ for the method of \cite{Zhang.2014b}.

Furthermore, we investigated the effect of training different models for different parts of the target pose vector. The results for three approaches with different label splitting are shown in Table~\ref{tab:6dvs3d}. For position prediction, splitting up the training improves performance. However, training on a single position output does not lead to improvement. For orientation prediction, removing the position part does not have a substantial effect. Splitting the labels up further even deteriorates performance. Based on these observations, we choose to train position and orientation separately.


\begin{table*}
	\centering
	\begin{tabular}{l l l l l l l}
		& \multicolumn{3}{c}{Position} & \multicolumn{3}{c}{Orientation} \\ \\ 
		& 6D Label & \textbf{3D Label} & 1D Label & 6D Label & \textbf{3D Label} & 1D Label \\ \hline \\
		MAE & $\SI{25.32 \pm 15.4}{\micro\metre}$ & \boldmath $\SI{14.89 \pm 9.3}{\micro\metre}$ & $\SI{15.88 \pm 12.6}{\micro\metre}$ & $\SI{0.099 \pm 0.056}{\degree}$ & \boldmath $\SI{0.096 \pm 0.072}{\degree}$ & $\SI{0.119 \pm 0.117}{\degree}$ \\
		rMAE & $0.029 \pm 0.024$ & \boldmath $0.018 \pm 0.014$ & $0.019 \pm 0.015$ & $0.0173 \pm  0.015$ & \boldmath $0.0168 \pm 0.016$ & $0.021 \pm 0.020$ \\
		aCC & $0.9991$ & \boldmath $0.9996$ & \boldmath $0.9996$ & \boldmath $0.9996$ & \boldmath $0.9996$ & $0.9993$  \\ 
	\end{tabular} \\
	\caption{MAE, relative MAE (with standard deviation) and average correlation coefficient for position and orientation prediction when training on position and orientation separately or simultaneously. 6D label refers to training with the entire pose as the network output. 3D label refers to training of two separate networks for position and orientation. 1D label refers to training of six networks on one part of the pose label each. Note, that the relative MAE and average correlation coefficient do not have a unit since they are relative measures. The best category is marked bold. We used the Inception3D model and marker B for this experiment.}
	\label{tab:6dvs3d}
\end{table*}

\subsection{2D Depth Information vs. 3D Volume Information}

\begin{table*}
	\centering
	\begin{tabular}{l l l l l l l l l l l l l l l}
		& \multicolumn{7}{c}{Position} & \multicolumn{7}{c}{Orientation} \\ \\ 
		& \textbf{Vol.} & M1 & M3 & D1 & D3 & MD & V. 2D & \textbf{Vol.} & M1 & M3 & D1 & D3 & MD & V. 2D \\ \hline \\
		MAE & $\boldsymbol{23.65}$ & $46.16$ & $81.67$ & $58.32$ & $224.9$ & $43.45$ & $28.84$ & $\boldsymbol{0.268}$ & $0.741$ & $0.755$ & $0.763$ & $0.828$ & $0.597$ & $0.290$ \\
		rMAE & $\boldsymbol{0.028}$ & $0.061$ & $0.089$ & $0.073$ & $0.182$ & $0.057$ &  $0.034$ & $\boldsymbol{0.047}$ & $0.129$ & $0.132$ & $0.133$ & $0.145$ & $0.104$ & $0.051$ \\
		aCC & $\boldsymbol{0.999}$ & $0.993$ & $0.988$ & $0.991$ & $0.956$ & $0.994$ &  $0.998$ & $\boldsymbol{0.998}$ & $0.982$ & $0.982$ & $0.976$ & $0.975$ & $0.988$ & $0.997$ \\ 
	\end{tabular} \\
	\caption{MAE, rMAE and aCC for position and orientation prediction for 2D representations with a 2D CNN in comparison to volumetric data with a 3D CNN. M1 refers to MIPs extracted from volumes along the $z$-direction. M3 additionally adds projections from the lateral $x$- and $y$- dimension as additional channels of the 2D input image. D refers to the normalized pixel value of the MIPs, i.e., it represents a depth map along the respective dimension. MD refers to a mixture of maximum intensities and depth values where the input image contains one channel for each representation for the MIP along the $z$-direction. V. 2D refers to the use of volume data with a 2D CNN where $3\times  3 \times 1$ kernels are used and neighboring slices are considered in the channel dimension. Position MAEs are given in $\si{\micro\metre}$ and rotation MAEs are given in $\si{\degree}$. The rMAE and aCC do not have units since they are relative measures. The best performing model is marked bold. All models are based on Inception3D, for the 2D cases, the third dimension of filters is omitted. Marker A was used for this experiment.}
	\label{tab:results}
\end{table*}

As a second step, we compare the accuracy when using 2D depth representations or full volumetric data for learning. The results are shown in Table~\ref{tab:results}. We used our Inception3D architecture for training. For the 2D representations, we removed the filter's third dimension, resulting in Inception2D. We conducted the experiment with marker A. This marker largely shows surface structures in OCT volumes. Therefore, 2D depths maps could be expected to contain a similar amount of information for learning.

Considering the comparison between 2D and 3D, the volumetric data representation that is used for training Inception3D clearly outperforms all 2D approaches. Note, that the 2D CNN version has a smaller capacity since filters only cover two dimensions. However, the 2D CNN was always able to reach a similar training error. This shows that insufficient capacity cannot be the reason for the performance difference but rather the representations used for learning. 

Out of all models with 2D filters, the model with volume inputs performs best. Here, volume data is processed in a z-slice-wise fashion with $3 \times 3 \times 1$ kernels while also taking neighboring slices into account.

Considering the difference between 2D representations, it is notable that a combination of depth and intensity information from a single MIP in $z$-direction performs best. Moreover, the single channel representations that only leverage information from the $z$ direction perform better than representations with additional $x$-$z$ and $y$-$z$ projections.

\subsection{Surface vs. Subsurface Structure}

\begin{table*}
	\centering
	\begin{tabular}{l l l l l }
		& \multicolumn{2}{c}{Position} & \multicolumn{2}{c}{Orientation} \\ \\ 
		& Marker A & \textbf{Marker B} & Marker A & \textbf{Marker B} \\ \hline \\
		MAE & $\SI{23.65 \pm 16.0}{\micro\metre}$ & \boldmath $\SI{14.89 \pm 9.3}{\micro\metre}$ & $\SI{0.268 \pm 0.22}{\degree}$ & \boldmath $\SI{0.096 \pm 0.072}{\degree}$ \\
		rMAE & $0.028 \pm 0.024$ & \boldmath $0.018 \pm 0.014$ & $0.047 \pm  0.052$ & \boldmath $0.0168\pm 0.016$ \\
		aCC & $0.9986$ & \boldmath $0.9996$ & $0.9975$ & \boldmath $0.9996$  \\ 
	\end{tabular} \\
	\caption{MAE, rMAE (with standard deviation) and aCC for position and orientation prediction for the marker with surface structure (A) compared to the marker with a depth structure (B). Note, that the rMAE and aCC do not have units since they are relative measures. The best category is marked bold. We used the Inception3D model for this experiment.}
	\label{tab:markercomp}
\end{table*}

\begin{figure}
	\centering
	\includegraphics[scale=1]{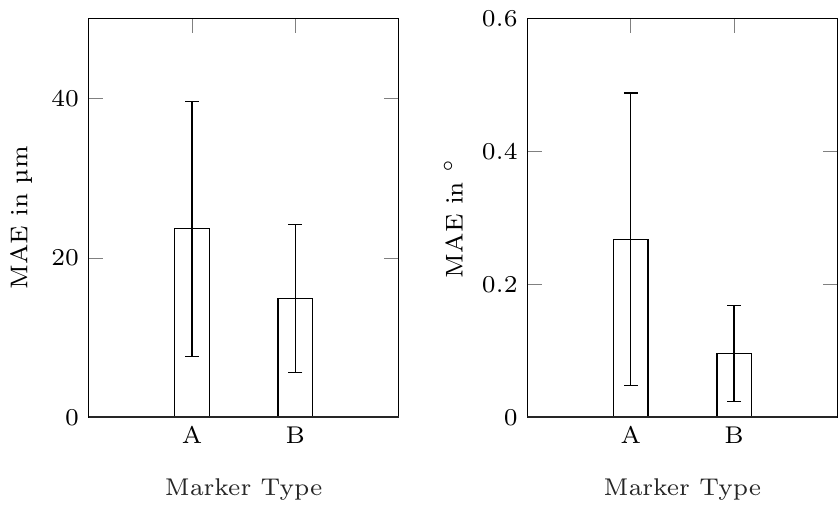}
	\caption{The comparison of test errors for a marker with surface structure (A) and a marker with a depth profile (B). Right, the rotation error in degree (MAE with standard deviation) is shown. Left, the position error in micrometer (MAE with standard deviation) is shown. We used the Inception3D model for this experiment.}
	\label{fig:markercomp}
\end{figure}

The last section compared a volumetric representation to 2D projections which are typically employed for 6D pose estimation frameworks. Next, we show how a recognizable inner structure affects learning for 3D CNNs. The two markers we compare are described in Section~\ref{sec:data}. Their key difference is that one marker has an opaque surface under infrared light (A), while the second marker has a visible inner structure in OCT images (B), see Figure~\ref{fig:markers}. The results are shown in Figure~\ref{fig:markercomp}. Detailed values are shown in Table~\ref{tab:markercomp}. Marker B clearly outperforms marker A. It is notable, that the position error goes beyond the assumed ground-truth label accuracy, induced by the robot's specified repeatability of $\SI{\pm 20}{\micro\metre}$. 

As a result, we show that a marker with a depth profile outperforms an opaque marker, which adds to the observation that volumetric representations outperform their 2D counterparts.

\subsection{Visualizing What was Learned}

Next, we aim for a deeper understanding of what was learned by the 3D CNN. In particular, we investigate whether the 3D CNN leveraged the depth information given in the second marker. We employ guided backpropagation to generate saliency maps for a test set image, see Section~\ref{sec:methods}. The saliency maps are generated by deriving the $3\times1$ output with respect to the input image. Thus, the final saliency maps we use can be interpreted as a gradient image which has the same size as the test image. Saliency maps indicate, which region in the image is largely responsible for the output, i.e., a change in that region leads to the largest change in the output. 

\begin{figure*}
	\centering
	\includegraphics[scale=0.5]{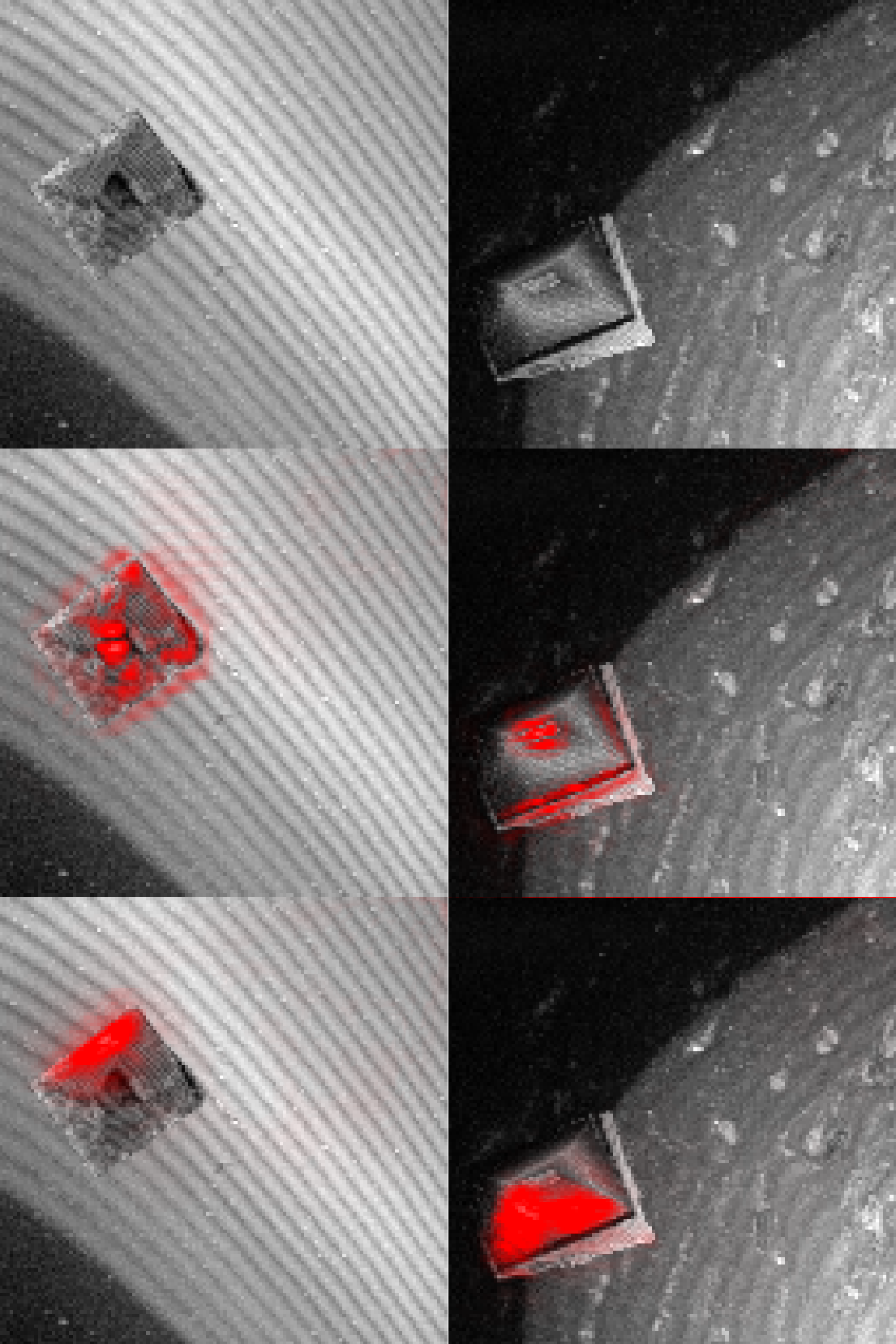}
	\caption{Comparison of saliency maps for the 2D and 3D data representations. Left, images of marker B are shown, right, images of marker A are shown. At the top, 2D MIPs along the axial $z$-direction of each marker in an OCT image are shown. In the middle, 2D saliency maps of the 2D training approach are shown in red, overlaid on the original input image. Here, the CNN (Inception2D) was directly trained on the 2D MIPs. At the bottom, 2D MIPs of the saliency maps are shown in red for the 3D CNN (Inception3D) that was trained on volume data. Here, MIPs of the volumetric saliency maps are overlaid on the input image's MIP. The saliency maps indicate which parts of the input image have the largest influence on the output. For 2D training, the saliency maps surround the marker's shape and focus on visible 2D features. For 3D training, the saliency maps do not appear to fit characteristic surface features.}
	\label{fig:sal2d}
\end{figure*}

To emphasize the importance of depth exploitation, we compare the 3D saliency maps from the two markers with 2D saliency maps from the approach of leveraging depth information from MIPs. The results for this are shown in Figure~\ref{fig:sal2d}. The saliency maps for the 2D CNN show high intensities at characteristic surface features on the markers. The 3D saliency maps for the 3D CNN, which are represented by 2D MIPs, focus on a region on the marker without sticking to specific surface features such as the pyramid tip. Note, that the same original test image was used for the 2D saliency maps and the 2D MIPs of the 3D saliency maps. 

\begin{figure*}
	\centering
	\includegraphics[scale=0.6]{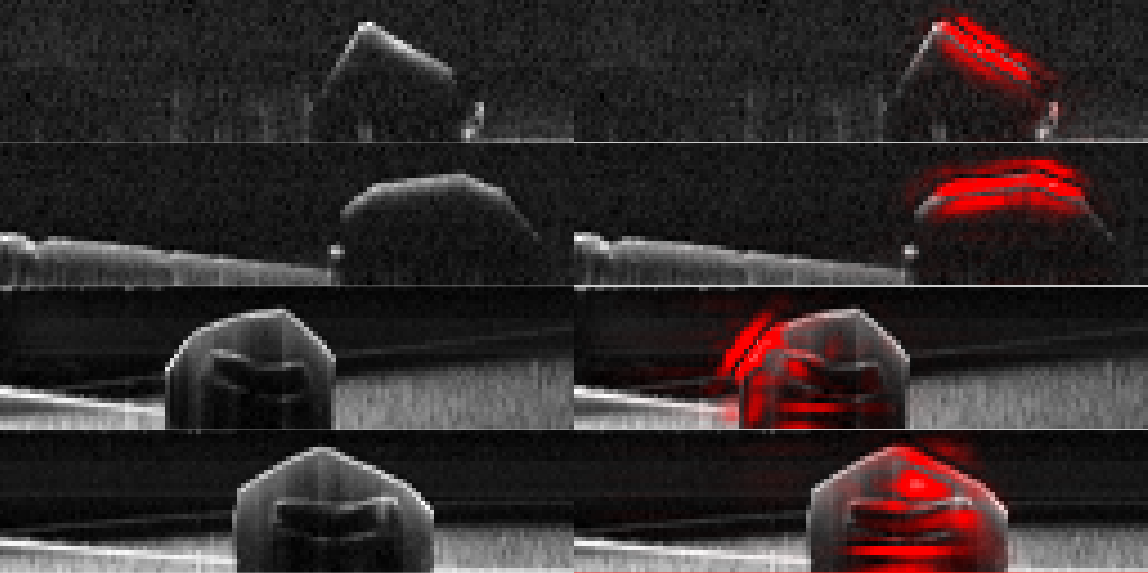}
	\caption{Visualization of what the 3D CNN focuses on using saliency maps. Left, two lateral slices through each marker are shown. The top two images show marker A, the bottom two show marker B. Right, slices through the 3D saliency maps for each marker are shown which are overlaid on the input image slices. The saliency maps show which region in the image shown on the left has the strongest influence on the CNN's output. The key difference between the markers' saliency maps is the focus on the marker's surface and inner structure, respectively. Both images and saliency maps are originally volumetric. Note, that the images were upsampled to twice their size from the 3D CNN input dimension of $64\times 64\times 16$.}
	\label{fig:salmarkers}
\end{figure*}

Furthermore, we present the saliency maps of two test images for the two markers in Figure~\ref{fig:salmarkers}. The saliency maps are shown in red as slices overlaid on top of slices of the test images. The cross-sectional view specifically shows what regions on and inside the marker have a large influence on the output. For the marker with a surface structure, the saliency map mostly lights up around the marker's surface. Note, that the high intensity saliency area spans above and below the surface, covering 3D space. For the marker with a depth structure, higher values in the saliency maps can be observed inside the marker. Furthermore, it should be noted that the 3D CNN's center of attention is indeed the marker itself. There appears to be no fitting on the ground surface or artifacts within the volume.

All in all, the visualization with saliency maps adds qualitative indications for depth exploitation of our 3D CNNs. This adds further insights to the quantitative results presented above.

\subsection{Inference Time and Robustness Towards Occlusion}

In this section, we show the applicability of our approach for practical problems. We provide results for the processing times of our CNNs to show that online pose estimation is feasible. Furthermore, we show results for our model when foreign objects appear in the OCT volume which is likely to happen in practice.


The results for inference time measurement are shown in Table~\ref{tab:inftimes}. We can observe that both CNNs allow sample processing at $\SI{50}{\hertz}$ with the 2D CNNs being slightly faster. Note, that the convolution operations only have a small influence with a total number of 68 out of 1734 operations and an average processing time of $\SI{0.065}{\milli\second}$ for Inception3D and $\SI{0.046}{\milli\second}$ for Inception2D. Also, note, that these values are very hardware and software dependent, see Section~\ref{sec:methods}.

\begin{table}
	\centering
	\begin{tabular}{l l}
		& Inference Time \\ \hline \\
		Inception3D & $\SI{20.95 +- 1.05}{\milli\second}$ \\
		Inception2D & $\SI{18.89 \pm 1.67}{\milli\second}$  \\ 
		Inception2D (Vol.) & $\SI{19.12 +- 1.59}{\milli\second}$  \\ 
	\end{tabular} \\
	\caption{Inference times (mean and standard deviation) for Inception3D, its 2D variant with 2D and 3D input data. The values are calculated based on 100 passes of a single sample through the network.}
	\label{tab:inftimes}
\end{table}

Furthermore, we investigate how well our model performs when the OCT volume is occluded with foreign objects, see Figure~\ref{fig:occlusion}. For this purpose, we use our third dataset where different objects are placed around the marker during data acquisition. The results are shown in Table~\ref{tab:occlusion}. The model's performance is still close to our other datasets where mostly the marker itself was visible. For rotations, the performance deteriorates more.

\begin{table}
	\centering
	\begin{tabular}{l l l}
		& Position & Orientation \\ \hline \\
		MAE & $\SI{16.62 \pm 8.4}{\micro\metre}$ & $\SI{0.187 \pm 0.093}{\degree}$  \\
		rMAE & $0.020 \pm 0.015$ & $0.040 \pm  0.032$ \\
		aCC & $0.9996$ &  $0.9988$\\ 
	\end{tabular} \\		
	\caption{MAE, rMAE (with standard deviation) and aCC for position and orientation prediction with our occlusion dataset, see Section~\ref{sec:methods} for a detailed description. Note, that the rMAE and aCC do not have units since they are relative measures. Marker B and Inception3D were used for this experiment.}
	\label{tab:occlusion}
\end{table}

\subsection{Architectures for Volumetric Data}

Next, we provide results on how different architecture designs behave for our pose estimation method. First, we present results for the four architectures introduced in Section~\ref{sec:methods}. Second, we show how long range feature propagation behaves for our Inception3D architecture.

\subsubsection{Comparison of 3D CNN Architectures}

\begin{table*}
	\centering
	\begin{tabular}{l l l l l l }
		& & \textbf{Inception3D} & ResNeXt3D & ResNetB3D & ResNetA3D \\ \hline \\
		\multirow{3}{*}{Marker A} & MAE & \boldmath $\SI{23.65 \pm 16.0}{\micro\metre}$ & $\SI{26.87 \pm 19.7}{\micro\metre}$ & $\SI{29.56 \pm 23.3}{\micro\metre}$ & $\SI{39.18 \pm 44.8}{\micro\metre}$ \\
		& rMAE & \boldmath $0.028 \pm 0.024$ & $0.031 \pm 0.028$ & $0.036 \pm 0.039$ & $0.044\pm 0.049$ \\
		& aCC & \boldmath $0.9986$ & $0.9984$ & $0.9973$ & $0.9962$  \\ \hline \\
		\multirow{3}{*}{Marker B} & MAE & \boldmath $\SI{14.89 \pm 9.3}{\micro\metre}$ & $\SI{16.28 \pm 10.6}{\micro\metre}$ & $\SI{17.68 \pm 11.0}{\micro\metre}$ & $\SI{21.71 \pm 11.7}{\micro\metre}$ \\
		& rMAE & \boldmath $0.018 \pm 0.014$ & $0.021 \pm 0.016$ & $0.022 \pm 0.018$ & $0.0275\pm 0.021$ \\
		& aCC & \boldmath $0.9996$ & $0.9994$ & $0.9993$ & $0.9991$  \\		
	\end{tabular} \\
	\caption{MAE, rMAE (with standard deviation) and aCC for position prediction with four different 3D CNN architectures, see Section~\ref{sec:methods} for a detailed description. Note, that the rMAE and aCC do not have units since they are relative measures. The best model is marked bold.}
	\label{tab:archres}
\end{table*}

For our deep learning framework, we propose four different models that come with different improved architectural ideas, see Section~\ref{sec:methods} for details. The results for position training are shown in Table~\ref{tab:archres}. With the most structural adjustments, Inception3D outperforms the other models. Furthermore, ResNetA3D, which uses the type of residual connections often employed for 3D CNNs \citep{Milletari.2016,Yu.2017b}, lacks behind more significantly.

\begin{figure}
	\centering
	\includegraphics[scale=1]{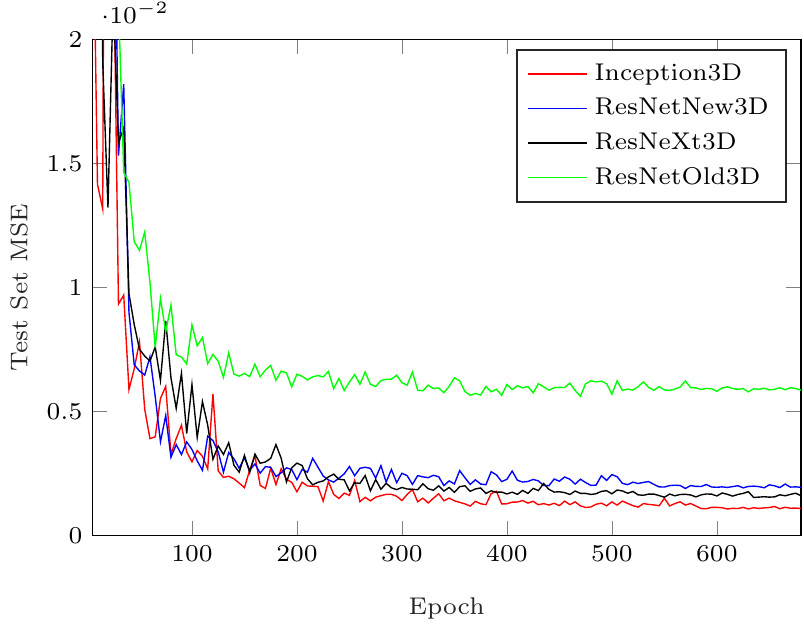}
	\caption{Comparison of test errors for all four architectures we introduce. The test set MSE during training is shown. Best viewed in color. The training behavior for the models being trained on marker A is shown.}
	\label{fig:archres}
\end{figure}

Additionally, Figure~\ref{fig:archres} shows the training behavior over time for all four models. In terms of convergence behavior, all models perform similar, as all models have approximately the same number of parameters. 

All in all, our results show improved performance for models that exploit more efficient architecture design principles.

\subsubsection{Long Range Residual Connections for Inception}

\begin{table*}
	\centering
	\begin{tabular}{l l l l l}
		& & \textbf{Residual} & Feature Based & None \\ \hline \\
		\multirow{3}{*}{Marker A} & MAE & \boldmath $\SI{23.65 \pm 16.0}{\micro\metre}$ & $\SI{23.99 \pm 17.2}{\micro\metre}$ & $\SI{27.17 \pm 22.3}{\micro\metre}$ \\
		& rMAE & \boldmath $0.028 \pm 0.024$ & $0.030 \pm 0.029$ & $0.033 \pm 0.039$ \\
		& aCC & \boldmath $0.9986$ & $0.9983$ & $0.9972$ \\ \hline \\
		\multirow{3}{*}{Marker B} & MAE & \boldmath $\SI{14.89 \pm 9.3}{\micro\metre}$ & $\SI{15.29 \pm 10.0}{\micro\metre}$ & $\SI{19.53 \pm 11.1}{\micro\metre}$ \\
		& rMAE & \boldmath $0.018 \pm 0.014$ & $0.021 \pm 0.016$ & $0.025 \pm 0.019$ \\
		& aCC & \boldmath $0.9996$ & $0.9994$ & $0.9992$ \\ 		
	\end{tabular} \\
	\caption{MAE, rMAE (with standard deviation) and aCC for position prediction with different types of long range connections, see Section~\ref{sec:methods} for a detailed description. Residual refers to long range residual connections, Feature refers to long range feature concatenation and None indicates no use of such connections. Note, that the rMAE and aCC do not have units since they are relative measures. The best model is marked bold.}
	\label{tab:resres}
\end{table*}

In the last section, we showed that our custom design of Inception3D outperforms other architectures. Next, we present results on how long range residual connections that span over modules affect performance. 

In Section~\ref{sec:methods} we presented two types of long range connections which are frequently used for feature transfer between similar sized stages in 3D CNNs for segmentation. We extend this approach by drawing connections between different stages of the network and introduce the concept to Inception3D by creating long range connections between modules. In Table~\ref{tab:resres} the results for the use of residual connections, feature connections and no connections at all are shown. Note, that the use of long- and short-range residual connections is also referred to as mixed residual connections \citep{Yu.2017b} and feature connections are also called dense connections \citep{Huang.2016}. Residual connections perform best, closely followed by feature connections. The model with no connections at all shows worse results. It should be noted that performance changes are small compared to using an entirely different architecture.

\begin{figure}
	\centering
	\includegraphics[scale=1]{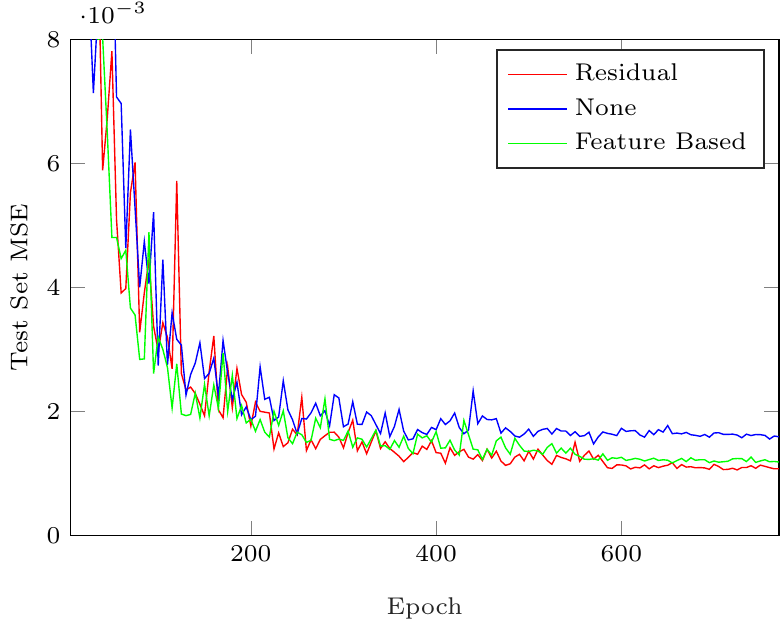}
	\caption{Comparison of test errors for three different scenarios regarding long range connections. Residual refers to long range residual connections, Feature refers to long range feature concatenation and None indicates no use of such connections. The test MSE during training is shown. Best viewed in color. The shown results are for training on marker A.}
	\label{fig:resres}
\end{figure}

In Figure~\ref{fig:resres} the training behavior of the three model variations is shown. There is a clear difference in errors for the model without any connections while the two models with connections are very close. The convergence behavior of the models is very similar once again. It should be noted that introducing the long range connections leads to a negligible increase in parameters.

Summarized, we showed various results highlighting the advantages of our novel deep learning-based pose estimation method. First, we showed that our method outperforms a comparable classic approach. Second, we showed that volumetric data leads to higher accuracy for pose learning, compared to depth-based approaches. Third, we provided qualitative saliency maps that demonstrate how 3D CNNs exploit inner features for pose estimation. Lastly, we showed results for our different architectures, highlighting the importance of efficient design principles with our proposed network Inception3D performing best.

\section{Discussion} \label{sec:discussion}

We provided extensive results for our method of 6D pose estimation from volumetric OCT data which lead to valuable insights for deep learning-based pose estimation and 3D CNN application to OCT in general.

\textbf{6D pose estimation from OCT volumes} with deep learning models is a novel approach. We motivate this idea by showing that we outperform other frameworks that rely on classical feature-based approaches \citep{Zhang.2014b}. This insight is in line with the general trend of deep learning methods replacing handcrafted features in many computer vision tasks \citep{Liefers.2017}.


Also, note, that position prediction accuracy is within the magnitude of the robot's repeatability and thus the ground-truth labels. Therefore, our deep learning approach is likely limited by the labels' accuracy and not a lack of representational power. In addition, our framework is general enough to be employed for various pose estimation problems as the source of labels can be any robot or motor.


Furthermore, we investigated how splitting up training for different parts of the pose affects performance with a significant improvement being observed when training only on positions, as shown in Table~\ref{tab:6dvs3d}. Often, multi-output regression is addressed by training a single model with multiple outputs instead of using multiple models with single outputs~\citep{Borchani.2015}. This approach promises better performance by introducing regularization through additional supervision. The model's feature maps have to learn to represent features for all outputs simultaneously. However, we observe performance improvement for position learning when splitting the pose label. This effect can be explained by regularization through learned invariance. When training on positions only, the input data contains examples with the marker being in the same position with different orientations. Thus, the CNN's weights are forced to learn invariance towards orientation. This is linked to OCT's properties as light scattering and surface visibility is highly dependent on the light beam's angle of impact. Therefore, invariance towards orientations also implicitly enforces invariance towards different light scattering properties in the data. Our results indicate, that the effect of learned invariance significantly improves position learning. At the same time, there are no significant performance differences for orientation learning. Shifting positions within the volume does not change the OCT's light beam angle of impact. Therefore, in opposite to position learning, invariance towards positions for rotation learning does not implicitly enforce invariance towards different light scattering conditions. All in all, our training strategy with split labels improves position learning by taking advantage of domain knowledge on OCT's light scattering properties.

\textbf{2D depth information and volume data} were investigated to draw a connection to OCT based tracking which has been performed on 2D projections \citep{Laves.2017}. The use of 2D depth representations can be motivated by the imaging property that many surfaces appear opaque under OCT as they cannot be penetrated by infrared light. Therefore, pure surface information extracted from the OCT volume could be deemed sufficient for most tasks. 

However, our results in Table~\ref{tab:results} show that moving towards volumetric data and 3D CNNs significantly increases performance. The use of volume data with flat 2D kernels already improves performance which indicates that a significant amount of information is lost when creating 2D projections. The novel approach of employing 3D CNNs for OCT volume data improves performance even further. The volumetric receptive fields of stacked 3D convolutional layers appear to be able to capture relevant features for pose estimation more effectively. 

With these findings we motivate the use of full volumetric information for OCT based tracking and pose estimation frameworks that relied on 2D representations so far \citep{Laves.2017,Camino.2016}. Other OCT based deep learning methods that have also relied on 2D representations so far \citep{Roth.2016,Wang.2016,Venhuizen.2015} could also benefit from our insights. 

We highlight the improved feature learning further with use of saliency maps for 2D and 3D data, see Figure~\ref{fig:sal2d}. For 2D data, the CNN appears to fit to distinct features on the marker surface that are visible in the 2D representation. The 3D CNN, however, appears to take advantage of other, deeper features that cannot be recognized on the surface. This leads to our investigation of deep subsurface feature learning.

\textbf{Markers with surface and subsurface structure} were compared to gain further insight on how 3D CNNs take advantage of inner features. Our results in Table~\ref{tab:markercomp} show that the marker with an inner structure performs significantly better than the marker that largely contains surface information in OCT images. This shows that the exploitation of OCT's 3D nature can be advantageous for volumetric feature learning with 3D CNNs. We support these quantitative result with additional saliency maps, see Figure~\ref{fig:salmarkers}. They highlight that the 3D CNNs indeed learned to exploit subsurface information when it was present in the volume data.

This finding shows that we can improve pose estimation performance without using a larger, more sophisticated marker. Ultimately, markers for surgery should be small and non-disruptive. Creating subsurface structures is an elegant solution to increase the learnable feature space without increasing the marker size. Thus, we combine the advantage of OCT's depth imaging with 3D CNN powered volumetric feature learning for pose estimation.

All in all, these insights emphasize once more, that OCT's capability of producing volumetric information is very exploitable by 3D CNNs. We provide strong evidence that OCT based 2D slicing and projection methods \citep{Roth.2016,Wang.2016,Venhuizen.2015} could significantly benefit from 3D data usage and volumetric feature exploitation.

\textbf{Moving towards clinical application scenarios} is a next step for our method. We highlight its suitability for future clinical use by showing its real-time processing capability and its robustness towards occlusion. 

Regarding the processing times shown in Table~\ref{tab:inftimes}, it is notable that the change between 2D and 3D convolutions does not lead to a significant difference. The largest processing overhead is caused by other operations that are always present in the network and neither the input size nor the different operations are a bottleneck. Therefore, our 3D CNNs are capable of online pose estimation. This is linked to our efficient 3D CNN architecture design with comparatively small numbers of parameters, as shown in Table~\ref{tab:allarch}.

For future application in clinical scenarios, our marker system should be capable of being integrated into existing OCT setups for MIS without requiring special operating conditions. Thus, it is crucial that our models deal well with unknown objects. Our occlusion dataset results in Table~\ref{tab:occlusion} show that our Inception3D model was able to learn robustness towards new occluding objects by achieving a performance close to the initial dataset.

The application of \textbf{deep learning architectures for 3D OCT data} is a novel approach. When entering new problem domains with the use of deep learning, it is largely unclear how existing models should be adopted \citep{Xie.2017}. Therefore, we created four different 3D CNN architectures with different design principles and showed how they affect performance for our novel learning problem.

In particular, the idea of downsampling intermediate network outputs with respect to their number of feature maps, i.e. creating a bottleneck, appears to improve representational power greatly. The only model without this property, ResNetA3D, performs significantly worse than the others, see Table~\ref{tab:archres}. The bottleneck idea has been successful for 2D CNNs \citep{He.2015} and we show that it is even more valuable for 3D CNNs. Bottlenecks address the key problem of model complexity and computaional cost which are particularily severe for 3D CNNs \citep{Yu.2017b}. The increased efficiency in terms of the number of parameters allows for much deeper models. This insight relates to \cite{Yu.2017} who built very deep 2D CNNs for medical image analysis by relying on downsampling in the feature map dimension.



In addition to the bottleneck principle, we use Inception3D and ResNeXt3D to address 3D CNN architecture design for our problem by showing the pay-off for extensive design and fine-tuning. Both architectures employ the successful principle of multiple paths at each scale \citep{Szegedy.2017}. However, for Inception3D, we carefully tuned each path individually while for ResNeXt3D, all paths are designed identically. Although there is a performance difference, it is notable that the simple design principles we followed for ResNeXt3D lead to a similar performance, see Table~\ref{tab:resres}. As a result, we argue that high-effort custom designs such as our Inception3D might not be strictly necessary for practice as more simple design choices can already reach good performance. Still, if the goal is the best performance possible, extensive fine-tuning will be necessary when entering new problem domains such as ours with 3D CNNs.


Additionally, we introduced long-range feature transfer between different scales for our architecture. This extends the idea of \cite{Ronneberger.2015} and \cite{Yu.2017b} who employed feature transfer between similar scales for segmentation tasks. As shown in Table~\ref{tab:resres}, these connections do lead to an improved performance. This supports the idea that we both need to detect our marker in the full image, which requires high level, coarse features with a large implicit FOV and we also need to detect fine grained differences for accurate pose distinction. The combination of fine, local and coarse, global features appears to lead to better pose estimation performance. This insight is in line with related ideas for object detection where features are also transferred for a combination of local and global properties \citep{Shrivastava.2016}.

Since the 3D CNN architectures we use are all very generic, our results have broader implications. In particular, it should be noted that the design principles of downsampling in the number of feature maps and multi-scale feature extraction are still rarely found in 3D medical image analysis. Early 3D CNN architectures have already been criticized for lack of representational capabilities \citep{Yu.2017b}. We extend on this point and argue that the design principles that we brought to the 3D domain with Inception3D and our other models are insufficiently applied for 3D medical learning problems. Several 3D CNN architectures with effective designs have been successfully introduced to the 3D image domain\cite{Chen.2017,Dou.2017,Kamnitsas.2017,Yu.2017b}. However, we argue that these well designed architectures could benefit further from the efficiency-focused design principles we introduced to 3D. Based on our results, we see a significant potential in current 2D CNN architectures for the 3D imaging domain. 

\section{Conclusions} \label{sec:conclusions}

We address the problem of high accuracy pose estimation for microscopic tracking tasks with OCT volume data. To this end, we introduce a novel deep learning-based pose estimation method that directly predicts a marker's pose from volumetric OCT data. We thoroughly analyze our method and compare to typical depth-based approaches which we convincingly outperform. Furthermore, 3D CNNs appear to exploit depth structures in volumetric data which we show both quantitatively with improved results and qualitatively with 3D saliency map visualizations. Our models are able to learn robustness towards occlusion which shows the markers' usability even when foreign objects appear in the OCT image which is likely to happen in a surgical scenario. Additionally, we show that efficient deep learning design principles can be effectively extended to the 3D image domain. Lastly, we showed that combining low- and high-level features through long range connections benefits pose learning.


For future work, OCT tracking frameworks could build on our insights and move towards deep learning based approaches with volume data exploitation. Furthermore, prior 2D based OCT learning approaches could be extended by volume based approaches. Regarding network architectures, future deep learning models for medical image analysis could incorporate more efficient architecture designs or directly adopt Inception3D for other problems.


\bibliography{jobname}

\begin{thebibliography}{69}
\expandafter\ifx\csname natexlab\endcsname\relax\def\natexlab#1{#1}\fi
\expandafter\ifx\csname url\endcsname\relax
  \def\url#1{\texttt{#1}}\fi
\expandafter\ifx\csname urlprefix\endcsname\relax\def\urlprefix{URL }\fi

\bibitem[{Abadi et~al.(2016)Abadi, Agarwal, Barham, Brevdo, Chen, Citro,
  Corrado, Davis, Dean, and Devin}]{Abadi.2016}
Abadi, M., Agarwal, A., Barham, P., Brevdo, E., Chen, Z., Citro, C., Corrado,
  G.~S., Davis, A., Dean, J., Devin, M., 2016. {Tensorflow: Large-scale machine
  learning on heterogeneous distributed systems}. {arXiv preprint
  arXiv:1603.04467}.

\bibitem[{Abdolmanafi et~al.(2017)Abdolmanafi, Duong, Dahdah, and
  Cheriet}]{Abdolmanafi.2017}
Abdolmanafi, A., Duong, L., Dahdah, N., Cheriet, F., 2017. {Deep feature
  learning for automatic tissue classification of coronary artery using optical
  coherence tomography}. {Biomedical Optics Express} 8~(2), 1203--1220.

\bibitem[{Allan et~al.(2014)Allan, Thompson, Clarkson, Ourselin, Hawkes, Kelly,
  and Stoyanov}]{Allan.2014}
Allan, M., Thompson, S., Clarkson, M.~J., Ourselin, S., Hawkes, D.~J., Kelly,
  J., Stoyanov, D., 2014. 2d-3d pose tracking of rigid instruments in minimally
  invasive surgery. In: International Conference on Information Processing in
  Computer-assisted Interventions. Springer, pp. 1--10.

\bibitem[{Borchani et~al.(2015)Borchani, Varando, Bielza, and
  Larra{\~n}aga}]{Borchani.2015}
Borchani, H., Varando, G., Bielza, C., Larra{\~n}aga, P., 2015. {A survey on
  multi-output regression}. {Wiley Interdisciplinary Reviews: Data Mining and
  Knowledge Discovery} 5~(5), 216--233.

\bibitem[{Bouget et~al.(2017)Bouget, Allan, Stoyanov, and Jannin}]{Bouget.2017}
Bouget, D., Allan, M., Stoyanov, D., Jannin, P., 2017. {Vision-based and
  marker-less surgical tool detection and tracking: a review of the
  literature}. {Medical Image Analysis} 35, 633--654.

\bibitem[{Brachmann et~al.(2014)Brachmann, Krull, Michel, Gumhold, Shotton, and
  Rother}]{Brachmann.2014}
Brachmann, E., Krull, A., Michel, F., Gumhold, S., Shotton, J., Rother, C.,
  2014. {Learning 6d object pose estimation using 3d object coordinates}. In:
  {European Conference on Computer Vision}. pp. 536--551.

\bibitem[{Brosch et~al.(2016)Brosch, Tang, Yoo, Li, Traboulsee, and
  Tam}]{Brosch.2016}
Brosch, T., Tang, L. Y.~W., Yoo, Y., Li, D. K.~B., Traboulsee, A., Tam, R.,
  2016. {Deep 3d convolutional encoder networks with shortcuts for multiscale
  feature integration applied to multiple sclerosis lesion segmentation}. {IEEE
  Transactions on Medical Imaging} 35~(5), 1229--1239.

\bibitem[{Camino et~al.(2016)Camino, Zhang, Gao, Hwang, Sharma, Wilson, Huang,
  and Jia}]{Camino.2016}
Camino, A., Zhang, M., Gao, S.~S., Hwang, T.~S., Sharma, U., Wilson, D.~J.,
  Huang, D., Jia, Y., 2016. Evaluation of artifact reduction in optical
  coherence tomography angiography with real-time tracking and motion
  correction technology. Biomedical optics express 7~(10), 3905--3915.

\bibitem[{Chen et~al.(2017{\natexlab{a}})Chen, Dou, Yu, Qin, and
  Heng}]{Chen.2017}
Chen, H., Dou, Q., Yu, L., Qin, J., Heng, P.-A., 2017{\natexlab{a}}.
  {VoxResNet: Deep voxelwise residual networks for brain segmentation from 3D
  MR images}. {NeuroImage}.

\bibitem[{Chen et~al.(2017{\natexlab{b}})Chen, Qi, Yu, Dou, Qin, and
  Heng}]{Chen.2017b}
Chen, H., Qi, X., Yu, L., Dou, Q., Qin, J., Heng, P.-A., 2017{\natexlab{b}}.
  {DCAN: Deep contour-aware networks for object instance segmentation from
  histology images}. {Medical Image Analysis} 36, 135--146.

\bibitem[{{\c{C}}i{\c{c}}ek et~al.(2016){\c{C}}i{\c{c}}ek, Abdulkadir,
  Lienkamp, Brox, and Ronneberger}]{Cicek.2016}
{\c{C}}i{\c{c}}ek, {\"O}., Abdulkadir, A., Lienkamp, S.~S., Brox, T.,
  Ronneberger, O., 2016. 3d u-net: learning dense volumetric segmentation from
  sparse annotation. In: International Conference on Medical Image Computing
  and Computer-Assisted Intervention. Springer, pp. 424--432.

\bibitem[{Dou et~al.(2016)Dou, Chen, Yu, Zhao, Qin, Wang, Mok, Shi, and
  Heng}]{Dou.2016b}
Dou, Q., Chen, H., Yu, L., Zhao, L., Qin, J., Wang, D., Mok, V. C.~T., Shi, L.,
  Heng, P.-A., 2016. {Automatic detection of cerebral microbleeds from MR
  images via 3D convolutional neural networks}. {IEEE Transactions on Medical
  Imaging} 35~(5), 1182--1195.

\bibitem[{Dou et~al.(2017)Dou, Yu, Chen, Jin, Yang, Qin, and Heng}]{Dou.2017}
Dou, Q., Yu, L., Chen, H., Jin, Y., Yang, X., Qin, J., Heng, P.-A., 2017. {3D
  deeply supervised network for automated segmentation of volumetric medical
  images}. {Medical Image Analysis}.

\bibitem[{Ehlers et~al.(2014)Ehlers, Srivastava, Feiler, Noonan, Rollins, and
  Tao}]{Ehlers.2014}
Ehlers, J.~P., Srivastava, S.~K., Feiler, D., Noonan, A.~I., Rollins, A.~M.,
  Tao, Y.~K., 2014. {Integrative advances for OCT-guided ophthalmic surgery and
  intraoperative OCT: Microscope integration, surgical instrumentation, and
  heads-up display surgeon feedback}. {PLoS One} 9~(8), e105224.

\bibitem[{Elfring et~al.(2010)Elfring, de~{La Fuente}, and
  Radermacher}]{Elfring.2010}
Elfring, R., de~{La Fuente}, M., Radermacher, K., 2010. {Assessment of optical
  localizer accuracy for computer aided surgery systems}. {Computer Aided
  Surgery} 15~(1-3), 1--12.

\bibitem[{Finke et~al.(2012)Finke, Kantelhardt, Schlaefer, Bruder, Lankenau,
  Giese, and Schweikard}]{Finke.2012}
Finke, M., Kantelhardt, S., Schlaefer, A., Bruder, R., Lankenau, E., Giese, A.,
  Schweikard, A., 2012. {Automatic scanning of large tissue areas in
  neurosurgery using optical coherence tomography}. {The International Journal
  of Medical Robotics and Computer Assisted Surgery} 8~(3), 327--336.

\bibitem[{Franz et~al.(2014)Franz, Haidegger, Birkfellner, Cleary, Peters, and
  Maier-Hein}]{Franz.2014}
Franz, A.~M., Haidegger, T., Birkfellner, W., Cleary, K., Peters, T.~M.,
  Maier-Hein, L., 2014. {Electromagnetic tracking in medicine---a review of
  technology, validation, and applications}. {IEEE Transactions on Medical
  Imaging} 33~(8), 1702--1725.

\bibitem[{Garc{\'\i}a-Peraza-Herrera et~al.(2016)Garc{\'\i}a-Peraza-Herrera,
  Li, Gruijthuijsen, Devreker, Attilakos, Deprest, Vander~Poorten, Stoyanov,
  Vercauteren, and Ourselin}]{GarciaPerazaHerrera.2016}
Garc{\'\i}a-Peraza-Herrera, L.~C., Li, W., Gruijthuijsen, C., Devreker, A.,
  Attilakos, G., Deprest, J., Vander~Poorten, E., Stoyanov, D., Vercauteren,
  T., Ourselin, S., 2016. Real-time segmentation of non-rigid surgical tools
  based on deep learning and tracking. In: International Workshop on
  Computer-Assisted and Robotic Endoscopy. Springer, pp. 84--95.

\bibitem[{Girshick et~al.(2014)Girshick, Donahue, Darrell, and
  Malik}]{Girshick.2014}
Girshick, R., Donahue, J., Darrell, T., Malik, J., 2014. Rich feature
  hierarchies for accurate object detection and semantic segmentation. In:
  Proceedings of the IEEE conference on computer vision and pattern
  recognition. pp. 580--587.

\bibitem[{Glorot et~al.(2011)Glorot, Bordes, and Bengio}]{Glorot.2011}
Glorot, X., Bordes, A., Bengio, Y., 2011. {Deep Sparse Rectifier Neural
  Networks}. In: {Aistats}. Vol.~15. p. 275.

\bibitem[{Greenspan et~al.(2016)Greenspan, {van Ginneken}, and
  Summers}]{Greenspan.2016}
Greenspan, H., {van Ginneken}, B., Summers, R.~M., 2016. {Guest editorial deep
  learning in medical imaging: Overview and future promise of an exciting new
  technique}. {IEEE Transactions on Medical Imaging} 35~(5), 1153--1159.

\bibitem[{Havaei et~al.(2017)Havaei, Davy, Warde-Farley, Biard, Courville,
  Bengio, Pal, Jodoin, and Larochelle}]{Havaei.2017}
Havaei, M., Davy, A., Warde-Farley, D., Biard, A., Courville, A., Bengio, Y.,
  Pal, C., Jodoin, P.-M., Larochelle, H., 2017. {Brain tumor segmentation with
  deep neural networks}. {Medical Image Analysis} 35, 18--31.

\bibitem[{He et~al.(2015)He, Zhang, Ren, and Sun}]{He.2015}
He, K., Zhang, X., Ren, S., Sun, J., 2015. {Delving deep into rectifiers:
  Surpassing human-level performance on imagenet classification}. In:
  {Proceedings of the IEEE International Conference on Computer Vision}. pp.
  1026--1034.

\bibitem[{He et~al.(2016)He, Zhang, Ren, and Sun}]{He.2016}
He, K., Zhang, X., Ren, S., Sun, J., 2016. {Deep residual learning for image
  recognition}. In: {Proceedings of the IEEE Conference on Computer Vision and
  Pattern Recognition}. pp. 770--778.

\bibitem[{Huang et~al.(2017)Huang, Liu, Weinberger, and van~der
  Maaten}]{Huang.2016}
Huang, G., Liu, Z., Weinberger, K.~Q., van~der Maaten, L., 2017. Densely
  connected convolutional networks. In: Proceedings of the IEEE conference on
  computer vision and pattern recognition. Vol.~1. p.~3.

\bibitem[{Ioffe and Szegedy(2015)}]{Ioffe.2015}
Ioffe, S., Szegedy, C., 2015. Batch normalization: Accelerating deep network
  training by reducing internal covariate shift. In: International conference
  on machine learning. pp. 448--456.

\bibitem[{Kamnitsas et~al.(2017)Kamnitsas, Ledig, Newcombe, Simpson, Kane,
  Menon, Rueckert, and Glocker}]{Kamnitsas.2017}
Kamnitsas, K., Ledig, C., Newcombe, V. F.~J., Simpson, J.~P., Kane, A.~D.,
  Menon, D.~K., Rueckert, D., Glocker, B., 2017. {Efficient multi-scale 3D CNN
  with fully connected CRF for accurate brain lesion segmentation}. {Medical
  Image Analysis} 36, 61--78.

\bibitem[{Karri et~al.(2017)Karri, Chakraborty, and Chatterjee}]{Karri.2017}
Karri, S.~P., Chakraborty, D., Chatterjee, J., 2017. {Transfer learning based
  classification of optical coherence tomography images with diabetic macular
  edema and dry age-related macular degeneration}. {Biomedical Optics Express}
  8~(2), 579--592.

\bibitem[{Kehl et~al.(2016)Kehl, Milletari, Tombari, Ilic, and
  Navab}]{Kehl.2016}
Kehl, W., Milletari, F., Tombari, F., Ilic, S., Navab, N., 2016. Deep learning
  of local rgb-d patches for 3d object detection and 6d pose estimation. In:
  European Conference on Computer Vision. Springer, pp. 205--220.

\bibitem[{Kingma and Ba(2014)}]{Kingma.2014}
Kingma, D., Ba, J., 2014. {Adam: A method for stochastic optimization}. In:
  {International Conference on Learning Representations}.

\bibitem[{Kral et~al.(2013)Kral, Puschban, Riechelmann, and
  Freysinger}]{Kral.2013}
Kral, F., Puschban, E.~J., Riechelmann, H., Freysinger, W., 2013. {Comparison
  of optical and electromagnetic tracking for navigated lateral skull base
  surgery}. {The International Journal of Medical Robotics and Computer
  Assisted Surgery} 9~(2), 247--252.

\bibitem[{Krizhevsky et~al.(2012)Krizhevsky, Sutskever, and
  Hinton}]{Krizhevsky.2012}
Krizhevsky, A., Sutskever, I., Hinton, G.~E., 2012. {ImageNet Classification
  with Deep Convolutional Neural Networks}. In: {F. Pereira}, {C. J. C.
  Burges}, {L. Bottou}, {K. Q. Weinberger} (Eds.), {Advances in Neural
  Information Processing Systems 25}. {Curran Associates, Inc}, pp. 1097--1105.

\bibitem[{Krull et~al.(2015)Krull, Brachmann, Michel, {Ying Yang}, Gumhold, and
  Rother}]{Krull.2015}
Krull, A., Brachmann, E., Michel, F., {Ying Yang}, M., Gumhold, S., Rother, C.,
  2015. {Learning Analysis-by-Synthesis for 6D Pose Estimation in RGB-D
  Images}. In: {The IEEE International Conference on Computer Vision (ICCV)}.

\bibitem[{Lankenau et~al.(2007)Lankenau, Klinger, Winter, Malik, M{\"u}ller,
  Oelckers, Pau, Just, and H{\"u}ttmann}]{Lankenau.2007}
Lankenau, E., Klinger, D., Winter, C., Malik, A., M{\"u}ller, H.~H., Oelckers,
  S., Pau, H.-W., Just, T., H{\"u}ttmann, G., 2007. {Combining optical
  coherence tomography (OCT) with an operating microscope}. In: {Advances in
  Medical Engineering}. Springer, pp. 343--348.

\bibitem[{Laves et~al.(2017)Laves, Schoob, Kahrs, Pfeiffer, Huber, and
  Ortmaier}]{Laves.2017}
Laves, M.-H., Schoob, A., Kahrs, L.~A., Pfeiffer, T., Huber, R., Ortmaier, T.,
  2017. {Feature tracking for automated volume of interest stabilization on
  4D-OCT images}. In: {SPIE Medical Imaging}. pp. 101350W--101350W.

\bibitem[{Lee et~al.(2017)Lee, Baughman, and Lee}]{Lee.2017}
Lee, C.~S., Baughman, D.~M., Lee, A.~Y., 2017. {Deep Learning Is Effective for
  Classifying Normal versus Age-Related Macular Degeneration Optical Coherence
  Tomography Images}. {Ophthalmology Retina}.

\bibitem[{Lee et~al.(2015)Lee, Xie, Gallagher, Zhang, and Tu}]{Lee.2015}
Lee, C.-Y., Xie, S., Gallagher, P., Zhang, Z., Tu, Z., 2015. {Deeply-supervised
  nets}. In: {Artificial Intelligence and Statistics}. pp. 562--570.

\bibitem[{Li et~al.(2017)Li, Zeng, and Ji}]{Li.2017}
Li, R., Zeng, T., Ji, S., 2017. {Deep Learning Segmentation of Optical
  Microscopy Images Improves 3D Neuron Reconstruction}. {IEEE Transactions on
  Medical Imaging}.

\bibitem[{Liefers et~al.(2017)Liefers, Venhuizen, Theelen, Hoyng, {van
  Ginneken}, and S{\'a}nchez}]{Liefers.2017}
Liefers, B., Venhuizen, F.~G., Theelen, T., Hoyng, C., {van Ginneken}, B.,
  S{\'a}nchez, C.~I., 2017. {Fovea Detection in Optical Coherence Tomography
  using Convolutional Neural Networks}. In: {SPIE Medical Imaging}. p. 1013302.

\bibitem[{Lin et~al.(2014)Lin, Chen, and Yan}]{Lin.2013}
Lin, M., Chen, Q., Yan, S., 2014. {Network in network}. In: {International
  Conference on Learning Representations}.

\bibitem[{Litjens et~al.(2017)Litjens, Kooi, Bejnordi, Setio, Ciompi,
  Ghafoorian, van~der Laak, van Ginneken, and S{\'a}nchez}]{Litjens.2017}
Litjens, G., Kooi, T., Bejnordi, B.~E., Setio, A. A.~A., Ciompi, F.,
  Ghafoorian, M., van~der Laak, J.~A., van Ginneken, B., S{\'a}nchez, C.~I.,
  2017. A survey on deep learning in medical image analysis. Medical image
  analysis 42, 60--88.

\bibitem[{Long et~al.(2015)Long, Shelhamer, and Darrell}]{Long.2015}
Long, J., Shelhamer, E., Darrell, T., 2015. {Fully convolutional networks for
  semantic segmentation}. In: {Proceedings of the IEEE Conference on Computer
  Vision and Pattern Recognition}. pp. 3431--3440.

\bibitem[{Milletari et~al.(2016)Milletari, Navab, and Ahmadi}]{Milletari.2016}
Milletari, F., Navab, N., Ahmadi, S.-A., 2016. {V-net: Fully convolutional
  neural networks for volumetric medical image segmentation}. In: {3D Vision
  (3DV), 2016 Fourth International Conference on}. pp. 565--571.

\bibitem[{Otte et~al.(2014)Otte, Otte, Wittig, H{\"u}ttmann, Kugler,
  Dr{\"o}mann, Zell, and Schlaefer}]{Otte.2014}
Otte, C., Otte, S., Wittig, L., H{\"u}ttmann, G., Kugler, C., Dr{\"o}mann, D.,
  Zell, A., Schlaefer, A., 2014. {Investigating Recurrent Neural Networks for
  OCT A-scan Based Tissue Analysis}. {Methods of Information in Medicine}
  53~(4), 245--249.

\bibitem[{Rajput et~al.(2016)Rajput, Antoni, Otte, Saathoff, Matth{\"a}us, and
  Schlaefer}]{Rajput.2016}
Rajput, O., Antoni, S.-T., Otte, C., Saathoff, T., Matth{\"a}us, L., Schlaefer,
  A., 2016. {High accuracy 3D data acquisition using co-registered OCT and
  kinect}. In: {Multisensor Fusion and Integration for Intelligent Systems
  (MFI), 2016 IEEE International Conference on}. pp. 32--37.

\bibitem[{Richter et~al.(2013)Richter, Trillenberg, Schweikard, and
  Schlaefer}]{Richter.2013}
Richter, L., Trillenberg, P., Schweikard, A., Schlaefer, A., 2013. {Stimulus
  intensity for hand held and robotic transcranial magnetic stimulation}.
  {Brain stimulation} 6~(3), 315--321.

\bibitem[{Ronneberger et~al.(2015)Ronneberger, Fischer, and
  Brox}]{Ronneberger.2015}
Ronneberger, O., Fischer, P., Brox, T., 2015. {U-net: Convolutional networks
  for biomedical image segmentation}. In: {International Conference on Medical
  Image Computing and Computer-Assisted Intervention}. pp. 234--241.

\bibitem[{Roth et~al.(2016)Roth, {Le Lu}, Liu, Yao, Seff, Cherry, Kim, and
  Summers}]{Roth.2016}
Roth, H.~R., {Le Lu}, Liu, J., Yao, J., Seff, A., Cherry, K., Kim, L., Summers,
  R.~M., 2016. {Improving Computer-Aided Detection Using Convolutional Neural
  Networks and Random View Aggregation}. {IEEE Transactions on Medical Imaging}
  35~(5), 1170--1181.

\bibitem[{Sahu et~al.(2016)Sahu, Mukhopadhyay, Szengel, and Zachow}]{Sahu.2016}
Sahu, M., Mukhopadhyay, A., Szengel, A., Zachow, S., 2016. {Tool and Phase
  recognition using contextual CNN features}. {arXiv preprint
  arXiv:1610.08854}.

\bibitem[{Sarikaya et~al.(2017)Sarikaya, Corso, and Guru}]{Sarikaya.2017}
Sarikaya, D., Corso, J., Guru, K., 2017. {Detection and Localization of Robotic
  Tools in Robot-Assisted Surgery Videos Using Deep Neural Networks for Region
  Proposal and Detection}. {IEEE Transactions on Medical Imaging}.

\bibitem[{Schlegl et~al.(2015)Schlegl, Glodan, Podkowinski, Waldstein,
  Gerendas, Schmidt-Erfurth, and Langs}]{Schlegl.2015}
Schlegl, T., Glodan, A.-M., Podkowinski, D., Waldstein, S.~M., Gerendas, B.~S.,
  Schmidt-Erfurth, U., Langs, G., 2015. {Automatic segmentation and
  classification of intraretinal cystoid fluid and subretinal fluid in 3D-OCT
  using convolutional neural networks}. {Investigative Ophthalmology {\&}
  Visual Science} 56~(7), 5920.

\bibitem[{Shrivastava et~al.(2016)Shrivastava, Sukthankar, Malik, and
  Gupta}]{Shrivastava.2016}
Shrivastava, A., Sukthankar, R., Malik, J., Gupta, A., 2016. {Beyond skip
  connections: Top-down modulation for object detection}. {arXiv preprint
  arXiv:1612.06851}.

\bibitem[{Simonyan et~al.(2014)Simonyan, Vedaldi, and
  Zisserman}]{Simonyan.2013}
Simonyan, K., Vedaldi, A., Zisserman, A., 2014. {Deep inside convolutional
  networks: Visualising image classification models and saliency maps}. In:
  {International Conference on Learning Representations}.

\bibitem[{Simonyan and Zisserman(2015)}]{Simonyan.2014}
Simonyan, K., Zisserman, A., 2015. {Very deep convolutional networks for
  large-scale image recognition}. In: {International Conference on Learning
  Representations}.

\bibitem[{Springenberg et~al.(2015)Springenberg, Dosovitskiy, Brox, and
  Riedmiller}]{Springenberg.2014}
Springenberg, J.~T., Dosovitskiy, A., Brox, T., Riedmiller, M., 2015. {Striving
  for simplicity: The all convolutional net}. In: {International Conference on
  Learning Representations}.

\bibitem[{Szegedy et~al.(2017{\natexlab{a}})Szegedy, Ioffe, Vanhoucke, and
  Alemi}]{Szegedy.2016}
Szegedy, C., Ioffe, S., Vanhoucke, V., Alemi, A.~A., 2017{\natexlab{a}}.
  Inception-v4, inception-resnet and the impact of residual connections on
  learning. In: AAAI. Vol.~4. p.~12.

\bibitem[{Szegedy et~al.(2017{\natexlab{b}})Szegedy, Ioffe, Vanhoucke, and
  Alemi}]{Szegedy.2017}
Szegedy, C., Ioffe, S., Vanhoucke, V., Alemi, A.~A., 2017{\natexlab{b}}.
  {Inception-v4, Inception-ResNet and the Impact of Residual Connections on
  Learning}. In: {AAAI}. pp. 4278--4284.

\bibitem[{Szegedy et~al.(2015)Szegedy, Liu, Jia, Sermanet, Reed, Anguelov,
  Erhan, Vanhoucke, and Rabinovich}]{Szegedy.2015}
Szegedy, C., Liu, W., Jia, Y., Sermanet, P., Reed, S., Anguelov, D., Erhan, D.,
  Vanhoucke, V., Rabinovich, A., 2015. {Going Deeper With Convolutions}. In:
  {Proceedings of the IEEE Conference on Computer Vision and Pattern
  Recognition}. pp. 1--9.

\bibitem[{Tao et~al.(2014)Tao, Srivastava, and Ehlers}]{Tao.2014}
Tao, Y.~K., Srivastava, S.~K., Ehlers, J.~P., 2014. {Microscope-integrated
  intraoperative OCT with electrically tunable focus and heads-up display for
  imaging of ophthalmic surgical maneuvers}. {Biomedical Optics Express} 5~(6),
  1877--1885.

\bibitem[{Toshev and Szegedy(2014)}]{Toshev.2014}
Toshev, A., Szegedy, C., 2014. {Deeppose: Human pose estimation via deep neural
  networks}. In: {Proceedings of the IEEE Conference on Computer Vision and
  Pattern Recognition}. pp. 1653--1660.

\bibitem[{Venhuizen et~al.(2015)Venhuizen, Grinsven, and
  Hoyng}]{Venhuizen.2015}
Venhuizen, F.~G., Grinsven, M., Hoyng, C.~B., 2015. {Vendor Independent Cyst
  Segmentation in Retinal SD-OCT Volumes using a Combination of Multiple Scale
  Convolutional Neural Networks}. {Medical Image Computing and Computer
  Assisted Intervention-Challenge on Retinal Cyst Segmentation}.

\bibitem[{Wang et~al.(2016)Wang, Zhang, Yao, Zhao, and Zhou}]{Wang.2016}
Wang, Y., Zhang, Y., Yao, Z., Zhao, R., Zhou, F., 2016. {Machine learning based
  detection of age-related macular degeneration (AMD) and diabetic macular
  edema (DME) from optical coherence tomography (OCT) images}. {Biomedical
  Optics Express} 7~(12), 4928--4940.

\bibitem[{Wohlhart and Lepetit(2015)}]{Wohlhart.2015}
Wohlhart, P., Lepetit, V., 2015. {Learning descriptors for object recognition
  and 3d pose estimation}. In: {Proceedings of the IEEE Conference on Computer
  Vision and Pattern Recognition}. pp. 3109--3118.

\bibitem[{Xie et~al.(2017)Xie, Girshick, Doll{\'a}r, Tu, and He}]{Xie.2017}
Xie, S., Girshick, R., Doll{\'a}r, P., Tu, Z., He, K., 2017. Aggregated
  residual transformations for deep neural networks. In: 2017 IEEE Conference
  on Computer Vision and Pattern Recognition (CVPR). IEEE, pp. 5987--5995.

\bibitem[{Yu et~al.(2017{\natexlab{a}})Yu, Chen, Dou, Qin, and Heng}]{Yu.2017}
Yu, L., Chen, H., Dou, Q., Qin, J., Heng, P.-A., 2017{\natexlab{a}}. {Automated
  melanoma recognition in dermoscopy images via very deep residual networks}.
  {IEEE Transactions on Medical Imaging} 36~(4), 994--1004.

\bibitem[{Yu et~al.(2017{\natexlab{b}})Yu, Yang, Chen, Qin, and
  Heng}]{Yu.2017b}
Yu, L., Yang, X., Chen, H., Qin, J., Heng, P.-A., 2017{\natexlab{b}}.
  {Volumetric ConvNets with Mixed Residual Connections for Automated Prostate
  Segmentation from 3D MR Images}. In: {AAAI}. pp. 66--72.

\bibitem[{Zagoruyko and Komodakis(2016)}]{Zagoruyko.2016}
Zagoruyko, S., Komodakis, N., 2016. {Wide residual networks}. {arXiv preprint
  arXiv:1605.07146}.

\bibitem[{Zeiler and Fergus(2014)}]{Zeiler.2014}
Zeiler, M.~D., Fergus, R., 2014. {Visualizing and understanding convolutional
  networks}. In: {European Conference on Computer Vision}. pp. 818--833.

\bibitem[{Zhang and W{\"o}rn(2014)}]{Zhang.2014b}
Zhang, Y., W{\"o}rn, H., 2014. {Optical coherence tomography as highly accurate
  optical tracking system}. In: {IEEE/ASME International Conference on Advanced
  Intelligent Mechatronics (AIM), 2014}. IEEE, Piscataway, NJ, pp. 1145--1150.

\end{thebibliography}

\end{document}